\documentclass{article}

\usepackage{microtype}
\usepackage{graphicx}
\usepackage{subfigure}
\usepackage{booktabs} 

\usepackage{hyperref}

\usepackage[accepted]{icml2025}

\usepackage{amsmath}
\usepackage{amssymb}
\usepackage{mathtools}
\usepackage{amsthm}

\usepackage[capitalize,noabbrev]{cleveref}

\usepackage{xcolor}         

\theoremstyle{plain}
\newtheorem{theorem}{Theorem}[section]
\newtheorem{prop}[theorem]{Proposition}

\theoremstyle{definition}

\theoremstyle{remark}

\usepackage[textsize=tiny]{todonotes}

\icmltitlerunning{Neural Data Similarity and Biologically Plausible Temporal Credit Assignment Rules}

\begin{document}

\twocolumn[
\icmltitle{Can Biologically Plausible Temporal Credit Assignment Rules Match BPTT for Neural Similarity? E-prop as an Example}



\icmlsetsymbol{equal}{*}

\begin{icmlauthorlist}
\icmlauthor{Yuhan Helena Liu}{csml,pni}
\icmlauthor{Guangyu Robert Yang}{mit,altera}
\icmlauthor{Christopher J. Cueva}{mit,cpsy}
\end{icmlauthorlist}

\icmlaffiliation{csml}{Center for Statistics and Machine Learning, Princeton University, Princeton, NJ, USA}
\icmlaffiliation{pni}{Princeton Neuroscience Institute, Princeton University, Princeton, NJ, USA}
\icmlaffiliation{altera}{Altera, San Francisco Bay Area, CA, USA}
\icmlaffiliation{mit}{McGovern Institute for Brain Research, Massachusetts Institute of Technology, Cambridge, MA, USA}
\icmlaffiliation{cpsy}{Department of Cognitive and Psychological Sciences, Brown University, Providence, RI, USA}

\icmlcorrespondingauthor{Yuhan Helena Liu}{hl7582@princeton.edu}
\icmlcorrespondingauthor{Christopher J. Cueva}{ccueva@gmail.com}

%
%

\icmlkeywords{Machine Learning, ICML}

\vskip 0.3in
]



\printAffiliationsAndNotice{} 

\begin{abstract}
Understanding how the brain learns may be informed by studying biologically plausible learning rules. These rules, often approximating gradient descent learning to respect biological constraints such as locality, must meet two critical criteria to be considered an appropriate brain model: (1) good neuroscience task performance and (2) alignment with neural recordings. While extensive research has assessed the first criterion, the second remains underexamined. Employing methods such as Procrustes analysis on well-known neuroscience datasets, this study demonstrates the existence of a biologically plausible learning rule --- namely e-prop, which is based on gradient truncation and has demonstrated versatility across a wide range of tasks --- that can achieve neural data similarity comparable to Backpropagation Through Time (BPTT) when matched for task accuracy. Our findings also reveal that model architecture and initial conditions can play a more significant role in determining neural similarity than the specific learning rule. Furthermore, we observe that BPTT-trained models and their biologically plausible counterparts exhibit similar dynamical properties at comparable accuracies. These results underscore the substantial progress made in developing biologically plausible learning rules, highlighting their potential to achieve both competitive task performance and neural data similarity.
\end{abstract}

\section{Introduction}

Understanding how animals learn complex behaviors that span multiple temporal scales is a fundamental question in neuroscience. Effectively updating synaptic weights to achieve such learning requires solving the temporal credit assignment problem: determining how to assign the contribution of past neural states to future outcomes. In pursuit of answers, neuroscientists have increasingly adopted the mathematical framework of training recurrent neural networks (RNNs) as a model for brain learning mechanisms, inspired by seminal works that laid the foundation for this approach~\cite{Zipser1991, Fetz1992, Moody1998, mante2013context}. This strategy has driven significant advancements in developing biologically plausible (bio-plausible) learning rules, which aim to model learning processes that respect biological constraints (e.g., locality~\cite{marschall2019evaluating,bredenberg2023formalizing}) while achieving good task performance~\cite{lillicrap2020backpropagation,richards2019deep}. Specifically, biological plausibility here refers to learning rules where all signals required for weight updates are physically available at the synapse~\cite{marschall2019evaluating}. Beyond synaptic-level constraints, a learning rule appropriate for the brain must also meet two key criteria: (1) it must support good task performance, and (2) it must produce brain-like activity. While extensive work has addressed the first question, the second --- whether bio-plausible learning rules yield brain-like representations --- remains underexamined. 

Navigating the vast space of computational models --- which vary not only in learning rules but also in architecture and tasks~\cite{richards2019deep,zador2019critique,yang2021next} --- necessitates a systematic comparison of model representations with empirical brain data. To address this challenge, a variety of methods have been developed, aiming to quantify the similarity between computational models and neural data. Among these, popular methodologies include linear regression~\cite{yamins2014performance}, Representational Similarity Analysis (RSA)~\cite{kriegeskorte2008representational}, Centered Kernel Alignment (CKA)~\cite{kornblith2019similarity}, Singular Vector Canonical Correlation Analysis~\cite{raghu2017svcca}, Procrustes distance~\cite{williams2021generalized,ding2021grounding,duong2022representational}, and Dynamical Similarity Analysis (DSA)~\cite{ostrow2024beyond}. 
By comparing the geometry of state representations or the dynamics of neural activity, these methods provide a critical framework for evaluating the extent to which models approximate neural systems. 

Leveraging existing comparison methodologies, we compute the similarity scores of RNN models trained with bio-plausible learning rules to experimental data. Specifically, we evaluate these similarity scores by comparing them to those achieved by Backpropagation Through Time (BPTT)-trained models, as BPTT-trained RNNs are the predominant method for brain modeling and the benchmark that many bio-plausible learning rules aim to approximate~\cite{yang2020artificial,richards2019deep,lillicrap2020backpropagation}. This comparison allows us to assess the efficacy of bio-plausible approximations of gradient-descent learning in capturing neural data similarity. \textbf{Has the pursuit of biologically plausible learning rules obeyed plausibility at the level of synaptic implementation (e.g., locality) at the expense of reproducing brain-like neural activity --- or can some such rules, under the right conditions, yield representations as brain-like as those learned by BPTT?}

\textbf{Main contributions:} Our findings reveal that the distance between data and models trained with truncation-based bio-plausible learning rules is comparable to the distance achieved by models trained using BPTT. We specifically focus on learning rules that approximate the gradient by truncating bio-implausible terms, as these truncation-based bio-plausible rules, such as e-prop, have demonstrated efficacy and versatility in learning non-trivial tasks~\cite{bellec2020solution,eyono2022current}. Other bio-plausible training strategies for RNNs either have limited versatility or success on non-trivial tasks (see Related Works). Specifically, our contributions include:
\begin{itemize}
    \item We demonstrate there \textit{exists} a biologically plausible learning rule --- e-prop, which is a local learning rule based on gradient truncation and versatile across tasks --- that can achieve neural data similarity comparable to BPTT at equal accuracies. We also examine ModProp in Appendix Figure~\ref{fig:var_cond}. This is achieved by benchmarking well-known neuroscience datasets --- Mante 2013~\cite{mante2013context} and Sussillo 2015~\cite{sussillo2015neural} --- which are primate datasets chosen for their higher task difficulty compared to datasets from other species, using state-of-the-art similarity methods, particularly Procrustes distance (Figure~\ref{fig:default_distances} and Appendix Figure~\ref{fig:var_cond}).
    
    \item Second, we show that factors such as architecture and initial conditions --- particularly initial weight settings --- can have a more pronounced impact on neural data similarity than the specific choice of learning rule. This highlights that variations in these factors can surpass differences observed across learning rules (Figure~\ref{fig:gain}).

    \item Third, to explain the comparable similarities between BPTT and e-prop, we investigate their commonalities. We show their increased similarity at a lower learning rate for BPTT (Figure~\ref{fig:lrsim}). We also analyze their resemblance in terms of their top demixed principal components (Figure~\ref{fig:dPCA}), post-training weight eigenspectrum, and dynamical properties, explored via Dynamical Similarity Analysis (DSA) in Appendix Figure~\ref{fig:W_eig}. 
\end{itemize} 

\section{Related Works} \label{scn:related_works}

Understanding the mechanisms through which the brain learns, utilizing its myriad elements, remains a perennial quest in neuroscience. Recent years have seen a resurgence of interest in proposing biologically plausible learning rules~\cite{lillicrap2020backpropagation, scellier2017equilibrium, hinton2022forward, laborieux2022holomorphic, greedy2022single, sacramento2018dendritic, payeur2020burst, roelfsema2018control, aljadeff2019cortical, meulemans2022least, murray2019local, bellec2020solution, liu2021cell, liu2022biologically, marschall2020unified, ghosh2023gradient,wang2024brainscale, confavreux2024meta, richards2019deep,kaleb2024feedback}, suggesting potential neural algorithms that leverage known biological components. Many of these rules are grounded in experimental observations of synaptic plasticity, including STDP~\cite{bi1998synaptic}, differential anti-Hebbian plasticity~\cite{xie1999spike}, burst-induced plasticity~\cite{remy2007dendritic}, and neuromodulated Hebbian rules shown to approximate backpropagation~\cite{payeur2020burst,aceituno2023learning}.

Despite these advances, relatively little research has examined whether such rules --- motivated by neuronal or synaptic-level constraints --- give rise to brain-like activity at the circuit or network level. A separate line of work focuses on inferring learning rules directly from neural recordings~\cite{nayebi2020identifying, ashwood2020inferring, lim2015inferring, kepple2021curriculum, portes2022distinguishing}, aiming to reconstruct the learning mechanisms underlying synaptic plasticity. In contrast, our work does not seek to identify the brain’s actual learning rule --- an especially difficult task when only post-learning data is available --- but instead asks: can an existing biologically plausible rule produce neural activity aligned with empirical recordings? Our framework evaluates the outcomes of learning, using flexible post hoc comparisons that do not rely on observing the learning process itself. While this approach does not provide direct mechanistic insight, it fills a critical gap by assessing whether existing bio-plausible rules can replicate the dynamics observed in the brain. Recent work has similarly begun comparing representations learned by different types of algorithms—for example,~\citet{codol2024brain} explore how brain-like neural dynamics for behavioral control can emerge through reinforcement learning—providing a complementary perspective to our approach. 

Our research focuses on learning rules for recurrent neural networks, which are extensively used in brain modeling~\cite{vyas2020computation,perich2021inferring,schuessler2020interplay,yang2019task,mante2013context,turner2021charting,valente2021probing,langdon2022latent,barak2017recurrent,song2016training,maheswaranathan2019universality,yang2021next}. This study specifically investigates local learning rules that truncate gradients, as these have shown promising results in task learning and offer versatility across various network architectures. A systematic review~\cite{marschall2020unified} recognized random feedback local online (RFLO) as the only fully local (hence bio-plausible) rule. Post-review developments include e-prop, an adaptation of RFLO for non-vanilla (particularly spike-based) RNNs~\cite{bellec2020solution}, and MDGL~\cite{liu2021cell} with its extension ModProp~\cite{liu2022biologically}, which further refine the gradient approximation by considering local modulatory signals~\cite{Smith2020}. These rules are notable for their effectiveness in bio-plausible temporal credit assignment, matching the performance of the more traditional BPTT in many settings~\cite{eyono2022current}. Our study will, therefore, concentrate on these specific learning rules due to their demonstrated efficacy and bio-plausibility. Further details of these rules are explained in Appendix~\ref{scn:learning_rules}. 

Alternative training strategies for RNNs exist, but they either face bio-plausibility issues, lack versatility across settings, or struggle to scale to complex tasks. For instance, equilibrium propagation and related rules depend on the equilibrium assumption~\cite{scellier2017equilibrium,meulemans2022least}; although there are architectures similar to equilibrium propagation or deep feedback control that do not make the equilibrium assumption~\cite{gilra2017predicting,kaleb2024feedback}. Within truncation-based methods, the SnAP-n algorithm introduced in \cite{Menick2020} allows customization by selecting the truncation level $n$. While SnAp-1 aligns closely with e-prop/RFLO, SnAp-2 and higher $n$ require storing a triple tensor, which poses $O(N^3)$ storage demands not yet proven feasible for neural circuits. Therefore, SnAp-n ($n \geq 2$) remains biologically implausible, while SnAp-1 effectively reduces to e-prop/RFLO under certain conditions. Beyond truncation, the KeRNL algorithm approximates long-term dependencies using first-order low-pass filters and updates parameters via node perturbation, yet this also challenges biological plausibility by requiring frequent meta-parameter updates. Other strategies like FORCE learning~\cite{sussillo2009generating} offer alternatives, but our scope assumes recurrent weight adjustment and the non-reservoir version faces issues with locality~\cite{marschall2019evaluating,bredenberg2023formalizing}. This study focuses on supervised learning, setting aside the broader field of reinforcement learning for future work, thus not covering certain learning rules like the one in~\cite{miconi2017biologically}.

Comparing high-dimensional neural responses across different systems and contexts is crucial in neuroscience~\cite{chung2021neural} for assessing model quality, determining invariant neural states, and aligning brain-machine interface recordings, among other tasks~\cite{schrimpf2018brain,chaudhuri2019intrinsic,degenhart2020stabilization,pagan2025individual}. Among the myriad of methods developed to quantify representational dissimilarity \cite{yamins2014performance,schrimpf2018brain,kriegeskorte2008representational,raghu2017svcca,shahbazi2021using,williams2021generalized,ding2021grounding,duong2022representational,khosla2023soft,lin2023topology,pospisil2023estimating,nejatbakhsh2024comparing} --- such as linear regression, Canonical Correlation Analysis (CCA), Centered Kernel Alignment (CKA), Representational Similarity Analysis (RSA), shape metrics, and Riemannian distance --- we focus on Procrustes distance for its ability to provide a proper metric for comparing the geometry of state representations, and because several weaknesses have been identified in other similarity measures that are, for example, biased due to high dimensionality, or may rely on low variance noise components of the data~\cite{kornblith2019similarity, Davari2022, Dujmovic2023, Elmoznino2024, cloos2024differentiable}. 
Additionally, we extend our investigation to include Dynamical Similarity Analysis (DSA~\cite{ostrow2024beyond}) in the Appendix, assessing system dynamics to complement our geometric analyses. Overall, the value of these existing measures stems from their ability to compare complex systems without fully understanding them by capturing key structures. However, this strength also poses a limitation: they focus on specific structures, and it remains uncertain whether these structures accurately capture the computational properties of interest. Therefore, developing new measures remains a crucial and intriguing endeavor~\cite{Sexton2022, sucholutsky2023getting, Lin2023, Klabunde2023, Bowers2023, Lampinen2024}.

\section{Preliminaries}

\subsection{RNN Training Setup}

Our RNN architecture consists of $N_{in}$ input units, $N$ hidden units, and $N_{out}$ readout units. The update mechanism for the hidden state at time $t$, $h_t \in \mathbb{R}^N$, follows the equation:
\begin{equation}
h_{t+1} = \beta h_t + (1-\beta) (W_h f(h_t) + W_x x_t),
\end{equation}
where $\beta = 1 - \frac{dt}{\tau_m} \in \mathbb{R}$ is the leak factor determined by the simulation time step $dt$ and membrane time constant $\tau_m$; $f(\cdot): \mathbb{R}^N \rightarrow \mathbb{R}^N$ represents the activation function (We used $retanh$ to mimic type-1 neuronal firing and avoid negative firing rates but also explored $ReLU$ activation, as shown in Appendix Figure~\ref{fig:var_cond}); $W_h \in \mathbb{R}^{N \times N}$ and $W_x \in \mathbb{R}^{N \times N_{in}}$ are the recurrent and input weight matrices, respectively; and $x_t \in \mathbb{R}^{N_{in}}$ is the input at time $t$. The readout, $\hat{y}_{t} \in \mathbb{R}^{N{out}}$, is calculated as a linear combination of the hidden state's activation, $f(h_t)$, with the readout weights $w \in \mathbb{R}^{N_{out} \times N}$.

To train this RNN for the specific tasks in the datasets, we used synthetic input and target output detailed in Appendix~\ref{scn:learning_rules}. Our objective is to minimize the scalar loss $L \in \mathbb{R}$. For loss minimization, we examine various learning rules, including BPTT (our benchmark) that computes the exact gradient, $\nabla_W L(W_h) \in \mathbb{R}^{N \times (N_{in} + N + N_{out})}$, as well as bio-plausible learning rules that apply approximate gradients, $\tilde \nabla_W L(W) \in \mathbb{R}^{N \times (N_{in} + N + N_{out})}$:
\begin{equation}
\Delta W = -\eta \nabla_{W} L(W),
\end{equation}
\begin{equation}
\widehat{\Delta W} = -\eta \tilde \nabla_W L(W),
\end{equation}
where $W = [W_x \quad W_h \quad w^T] \in \mathbb{R}^{N \times (N_{in} + N + N_{out})}$ and $\eta \in \mathbb{R}$ is the learning rate. 

The learning rules investigated in this study are elaborated upon in Appendix~\ref{scn:learning_rules}. Our analysis centers on how training RNNs with different algorithms influences their similarity to neural data. Predominantly, we concentrate on the truncation-based, bio-plausible rule known as e-prop~\cite{bellec2020solution}, which simplifies the gradient by retaining only those terms that align with a three-factor learning rule. This includes a Hebbian eligibility trace modulated by a top-down instructive factor, potentially attributable to neuromodulators~\cite{Magee2020,Gerstner2018}. It is noteworthy that e-prop is equivalent to the RFLO learning rule introduced in~\cite{murray2019local} under most conditions. Additionally, we explore ModProp~\cite{liu2022biologically}, which incorporates cell-type-specific local modulatory signals~\cite{Smith2020} to recover terms omitted by e-prop. However, due to ModProp's limitations (constrained to the adherence of Dale's law and employing the $ReLU$ activation function), our examination of this rule is restricted to such specific contexts in Appendix Figure~\ref{fig:var_cond}.

\subsection{Similarity Measures}

As mentioned in the Introduction, we utilize the metric Procrustes distance~\cite{williams2021generalized} to quantify the similarity between the hidden states of RNN models, denoted by $H \in \mathbb{R}^{B * T \times N}$, and the experimentally recorded neural responses, represented as $\tilde{H} \in \mathbb{R}^{B * T \times N'}$. Here, $B$ represents the number of trials or experimental conditions, $T$ denotes the number of time steps in each trial, and $N$ and $N'$ correspond to the number of RNN hidden units and recorded neurons, respectively. The metric Procrustes distance can be viewed as the residual distance after the two neural representations are aligned with an optimal rotation, and is quantified as 
\begin{align}
\theta(H, \tilde{H}) &= \underset{Q \in \mathcal{O}}{\min} \arccos \left( \frac{<H^{\phi}, \tilde{H}^{\phi} Q>} {\| H^{\phi} \| \| \tilde{H}^{\phi} \|} \right)
\end{align}
where $\mathcal{O}$ is the group of orthogonal linear transformations~\cite{ding2021grounding, Harvey2023}. 

\section{Results} 

\begin{figure*}[h!]
    \centering
    \includegraphics[width=0.99\textwidth]{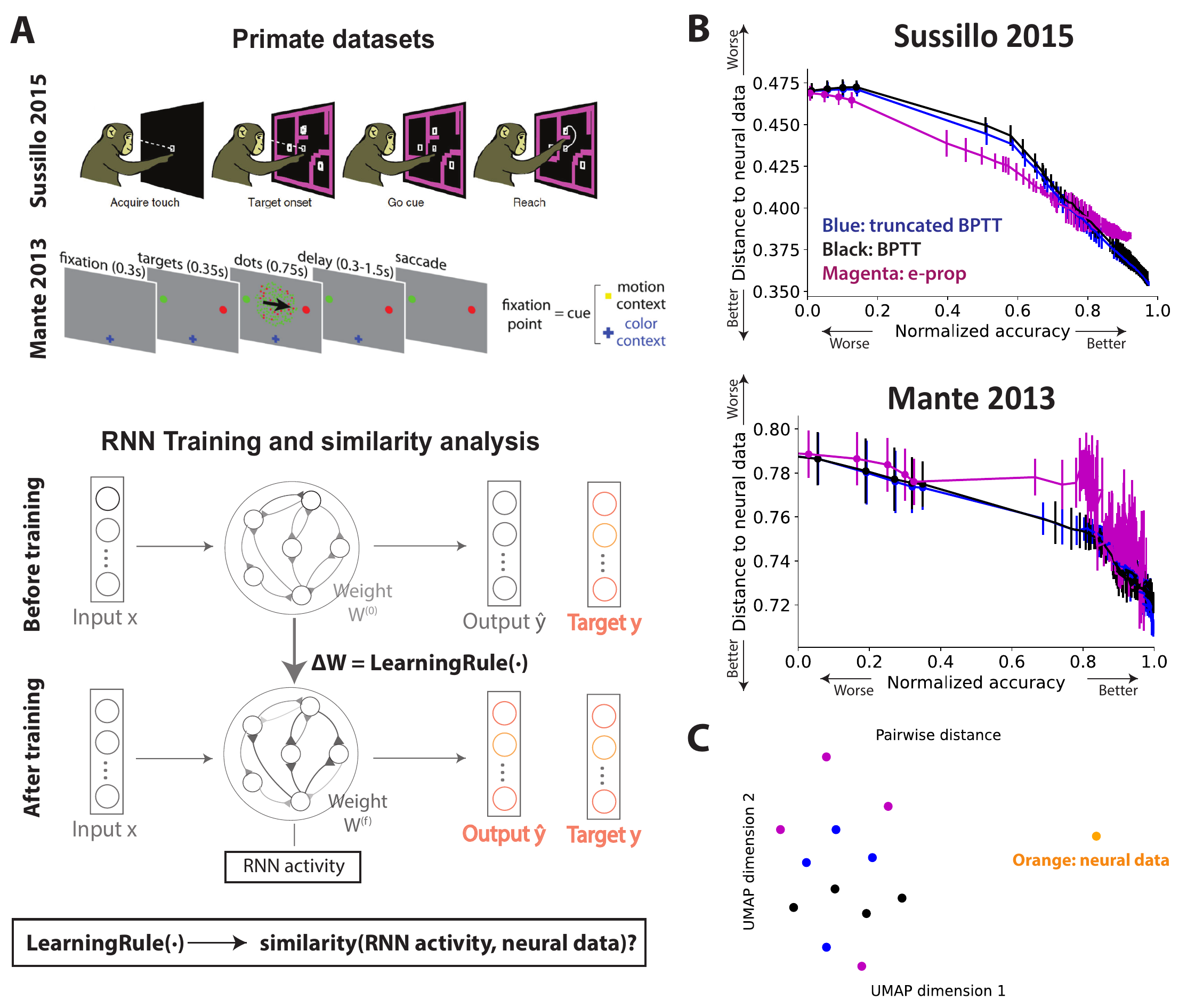}
    \caption{ (A) Setup overview: analysis of two neural datasets. We computed similarity scores between RNN activity and electrode recordings from (1)~\citet{mante2013context} and (2)~\citet{sussillo2015neural}. Schematics have been modified from those in the original papers. RNNs are trained on these respective tasks using various learning rules, including BPTT and bio-plausible alternatives. Subsequently, we evaluate the similarity between RNN activity post-training and the neural recordings to compare model-data similarity across different learning rules. (B) The Procrustes distance vs. accuracy plots for the Sussillo 2015 (top) and Mante 2013 (bottom) tasks illustrate that multiple learning rules achieve comparable data similarity. At each training iteration, both task accuracy and neural similarity are computed, and the resulting pairs are plotted—showing the progression of models during training 
    from low accuracy/high distance (upper left) to high accuracy/low distance (bottom right). This procedure is applied uniformly across all learning rules. Here, magenta denotes e-prop, blue denotes truncated BPTT (truncation length $=10$), and black denotes BPTT. The mean is plotted with error bars denoting standard deviations across four random seeds. The x-axis, normalized accuracy, is defined in Appendix~\ref{scn:sim_details}. (C) UMAP embedding of trained model representations with Sussillo 2015 as an illustrative example; additional training snapshots are shown in Appendix Figure~\ref{fig:pairDist_embedding_snapshot}. Additional similarity measures are given in Table~\ref{tab:additional_measures}. 
    } 
    \label{fig:default_distances}
\end{figure*} 

\begin{figure*}[h!]
    \centering
    \includegraphics[width=0.9\textwidth]{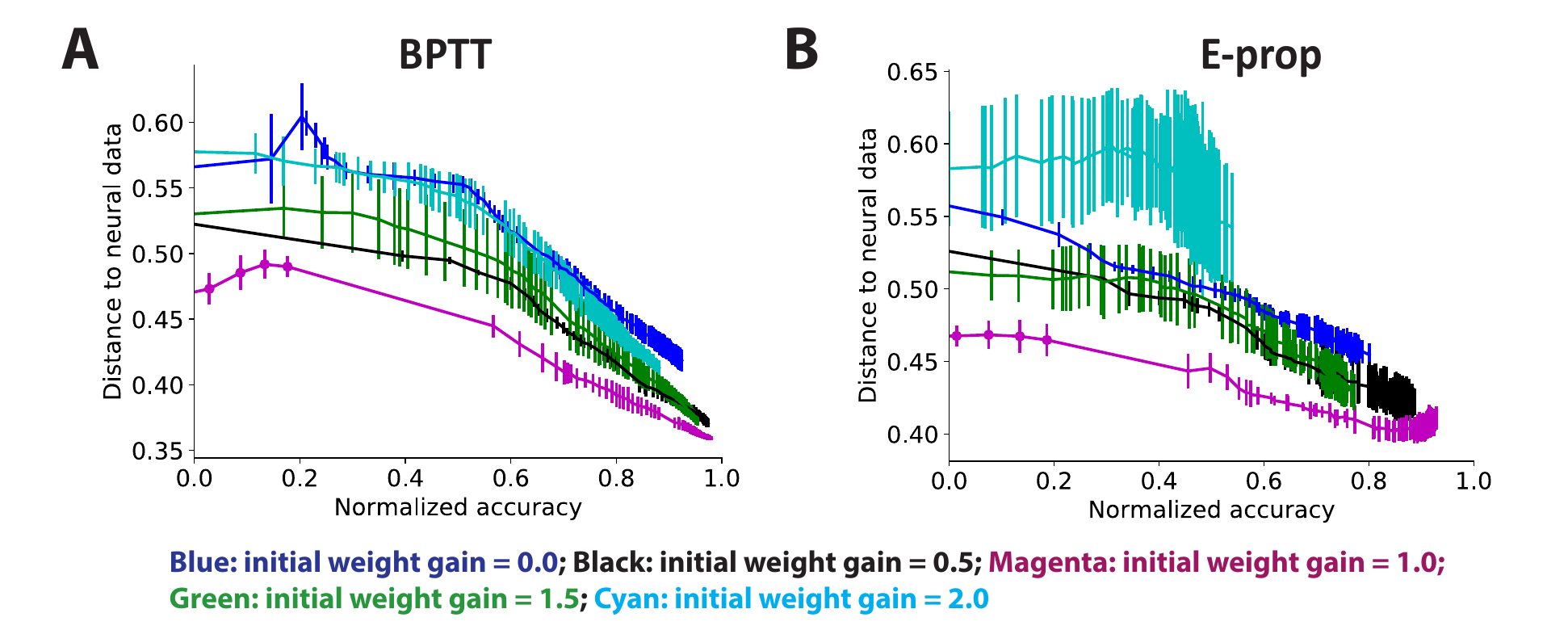}
    \caption{ \textbf{Impact of Initial Weight Magnitude on Model-Data Distances Exceeds Variation from Learning Rules}. Model-data distances versus normalized accuracy for various initial gain values (depicted by different colors) for (A) BPTT and (B) e-prop. Initial weight gain refers to the multiplier applied to the default initializations for recurrent and readout weights. The results shown are for the Sussillo 2015 task, with similar trends observed for the Mante 2013 task. The mean is plotted with error bars representing the standard deviation. Table~\ref{tab:procrustes_gains} further illustrates the significant influence of gain on distances.
    } 
    \label{fig:gain}
\end{figure*} 

In our study, we analyze the similarity between task-trained RNN models and two neural datasets: Sussillo 2015 \cite{sussillo2015neural} and Mante 2013 \cite{mante2013context}. An overview of our pipeline is provided in Figure~\ref{fig:default_distances}A, with detailed information about our RNN model setup, similarity measure, and datasets in Appendix~\ref{scn:methods}. We examine the similarity of RNN models, across different learning rules, to neural data, leveraging Procrustes analysis. Figure \ref{fig:default_distances}B shows that multiple learning rules, specifically BPTT and its truncation-based biologically plausible alternative (e-prop), achieve similar Procrustes distances from neural data across two distinct tasks: Sussillo 2015~\cite{sussillo2015neural} and Mante 2013~\cite{mante2013context}. Although the error bars for BPTT and e-prop do not appear to overlap near perfect accuracy in the Sussillo 2015 task, we demonstrate that such differences are minimal compared to other potential confounding factors in the brain, as shown in Figure~\ref{fig:gain} and Appendix Figure~\ref{fig:var_cond}. We further demonstrate the key trend using additional similarity measures in Table~\ref{tab:additional_measures}. 

Also, to see if the progress over time in developing more biologically plausible learning rules has led to improvements in aligning models with neural activity, we also evaluated older learning methods such as node perturbation and evolutionary strategies. Results show that these methods resulted in greater Procrustes distances compared to the aforementioned rules at equivalent accuracy levels, demonstrating that not all learning rules are equally effective. This also indicates the effectiveness of newer bio-plausible gradient-approximating learning rules over some of the older methods (Appendix Figure~\ref{fig:nodeP}).

\begin{table*}[ht]
\centering
\resizebox{0.99\textwidth}{!}{%
\begin{tabular}{lcccccc}
\toprule
\textbf{Rule} & \textbf{gain=0.0} & \textbf{gain=0.5} & \textbf{gain=1.0} & \textbf{gain=1.5} & \textbf{1.0 vs 0.0} & \textbf{1.0 vs 0.5} \\
\midrule
\textit{BPTT} & 0.461~$\pm$~0.006 & 0.423~$\pm$~0.005 & 0.398~$\pm$~0.007 & 0.428~$\pm$~0.010 & p=2.07e-5 & p=1.94e-3 \\
\textit{e-prop} & 0.461~$\pm$~0.007 & 0.437~$\pm$~0.009 & 0.407~$\pm$~0.008 & 0.432~$\pm$~0.008 & p=1.30e-4 & p=5.71e-3 \\
\textit{e-prop vs BPTT} & p=0.985 & p=0.078 & p=0.173 & p=0.587 & -- & -- \\
\textit{Noise ceiling} & 0.565~$\pm$~0.002 & 0.529~$\pm$~0.005 & 0.467~$\pm$~0.005 & 0.508~$\pm$~0.017 & -- & -- \\
\bottomrule
\end{tabular}
}
\caption{Procrustes distance across gains and learning rules for Figure~\ref{fig:gain}. Distances measured at normalized accuracy $\approx 0.8$. Values are mean~$\pm$~std across seeds. Noise ceiling corresponds to untrained models. Non-overlapping error bars across gains suggest gain influences neural distances; overlapping bars within columns suggest similarity across rules (for fixed gain). P-values (uncorrected t-tests due to only a small number of comparisons) highlight significant gain-dependent differences but insignificant differences across rules.}
\label{tab:procrustes_gains}
\end{table*}

Additionally, Figure \ref{fig:gain} delves into the impact of initial weight settings on model-data distances, revealing that such initial condition nuances exert a more pronounced influence than the choice of learning rule itself. Initial weight gain is a crucial attribute, as it significantly affects the dynamical properties of RNNs, particularly the Lyapunov exponents that govern the rates of expansion and contraction. It can also interpolate between rich and lazy learning regimes, imparting distinct inductive biases~\cite{braun2022exact,flesch2023continual,chizat2019lazy,woodworth2020kernel,schuessler2023aligned,bordelon2022influence,liu2023connectivity,paccolat2021geometric}. This finding further underscores the significant role of model initialization in shaping learning outcomes, with particular initial conditions facilitating a closer approximation to neural data than others. 

\begin{table}[ht]
\centering
\begin{tabular}{lcc}
\toprule
 & \textbf{CCA} & \textbf{CKA} \\
\midrule
BPTT & 0.272~$\pm$~0.009 & 0.152~$\pm$~0.015  \\
e-prop & 0.273~$\pm$~0.009 & 0.154~$\pm$~0.006 \\
p-val & p=0.953 & p=0.817  \\
\bottomrule
\end{tabular}
\caption{Additional similarity measures yield trends consistent with those in Figure~\ref{fig:default_distances}. We note that Sussillo 2015 is used here as an example. All p-values (uncorrected, due to the small number of comparisons) show no significant differences between trained BPTT and e-prop models across metrics (at matched accuracies).}
\label{tab:additional_measures} 
\end{table}

\begin{figure*}[h!]
    \centering
    \includegraphics[width=0.9\textwidth]{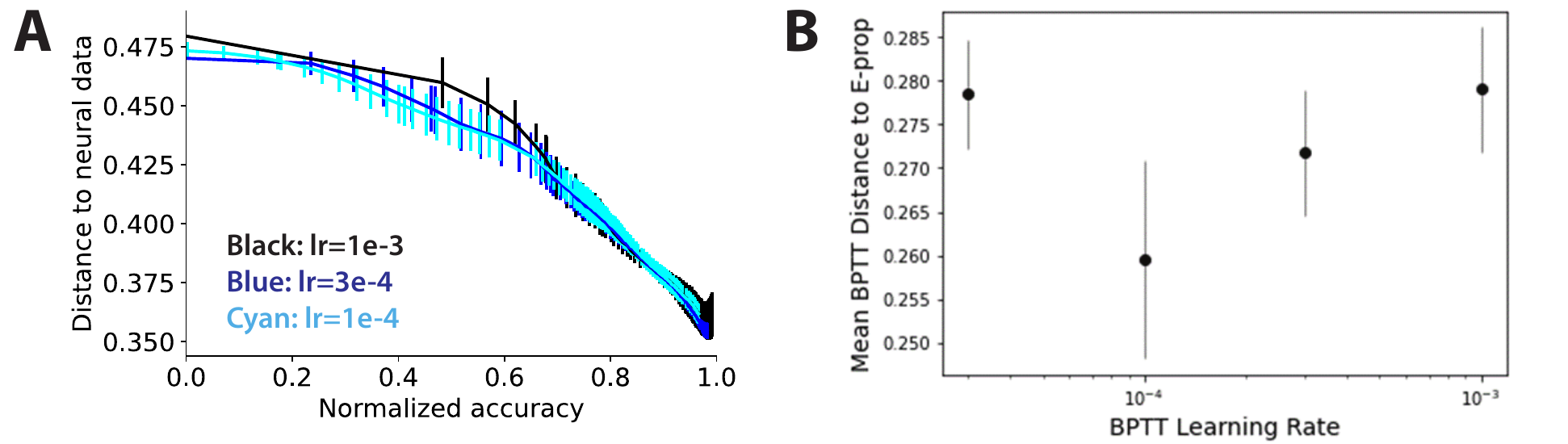}
    \caption{ (A) Procrustes distances remain consistent across various learning rates when employing the same rule (BPTT). Different color shades represent different learning rates: $1e-3$, $3e-4$, and $1e-4$. These rates result in nearly indistinguishable Procrustes distances. The analysis in this figure is done using the Sussillo 2015 task. (B) E-prop --- has been viewed as BPTT with a reduced learning rate plus some degree of gradient approximation error~\cite{liu2022beyond} --- aligns more closely with BPTT at a lower learning rate ($1e-4$) compared to the default setting ($1e-3$). Here, the mean distance from BPTT to e-prop is plotted, with error bars denoting the standard deviation.   
    } 
    \label{fig:lrsim}
\end{figure*}

\begin{figure*}[h!]
    \centering
    \includegraphics[width=0.9\textwidth]{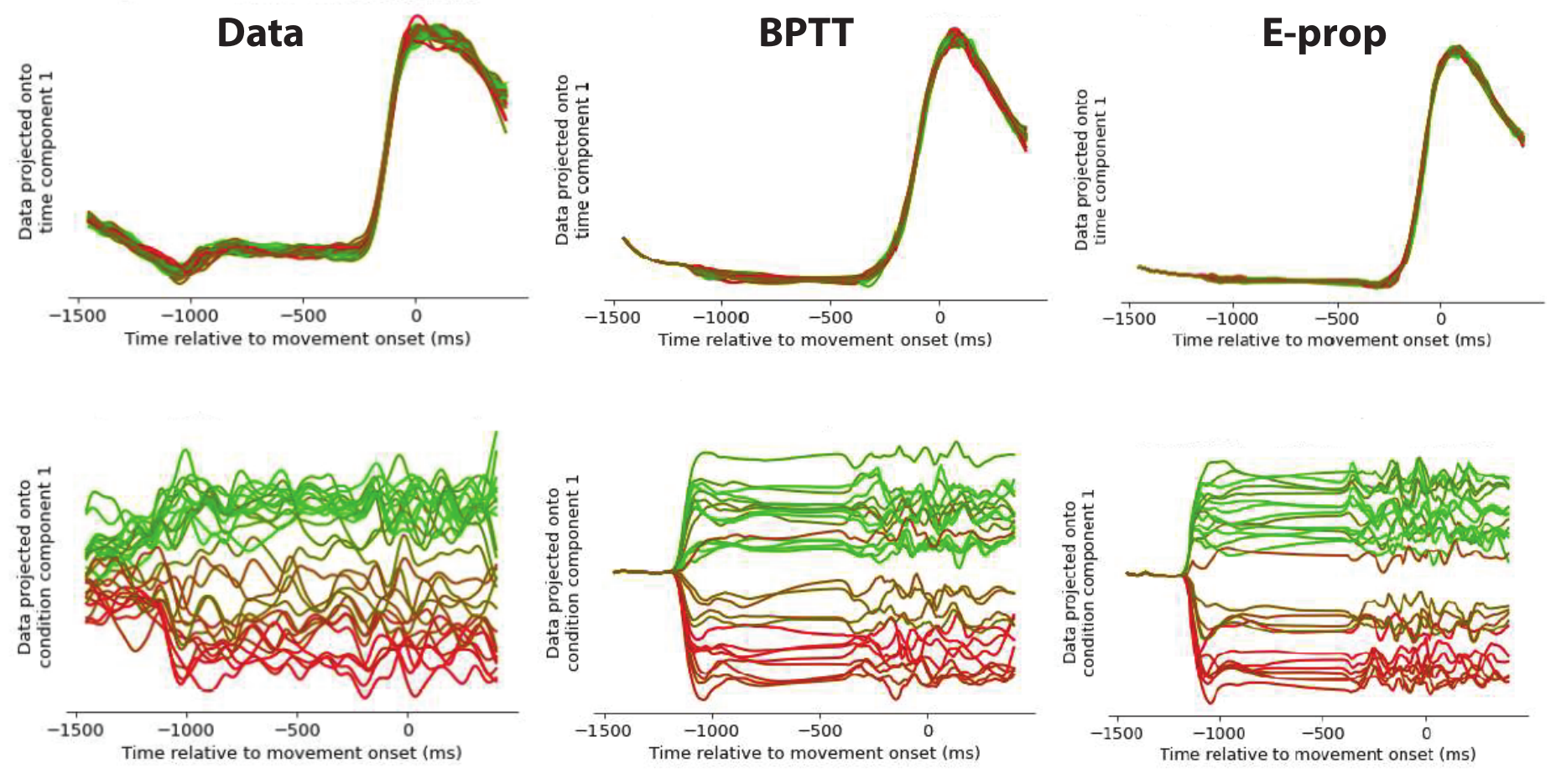}
    \caption{ Demixed principal component analysis (dPCA) shows a qualitative match between model and data when projected onto the first time component and first condition component. The time component captures variance aligned with task time, while the condition component captures variance attributable to input condition differences. Here, the Sussillo 2015 dataset is illustrated; similar trends are observed for the Mante 2013 dataset. Each color represents a different reach condition. 
    } 
    \label{fig:dPCA}
\end{figure*}

Figure \ref{fig:lrsim} explores the impact of learning rates on model-data distances across learning rules. In Figure \ref{fig:lrsim}A, Procrustes distances remain consistent across learning rates for BPTT. Given that e-prop can be decomposed into a lower learning rate BPTT and an approximation error~\cite{liu2022beyond}, which is further illustrated here by the similarity between a lower learning rate BPTT and e-prop (Figure \ref{fig:lrsim}B), this shared component of a lower learning rate BPTT could partly explain their similar distances. Additionally, post-training weight eigenspectrums and distances, analyzed via Dynamical Similarity Analysis (DSA), further reinforce the similarity between BPTT and e-prop (Appendix Figure~\ref{fig:W_eig}). This similarity is further explored in Figure~\ref{fig:dPCA}, where top demixed principal components show a qualitative match between the neural data and the models. We also display the similarity among models in terms of their pairwise distances and their embeddings across different sampled training snapshots in Appendix Figure~\ref{fig:pairDist_embedding_snapshot}. 

Building on the observed similarities between BPTT and e-prop, we derive Proposition~\ref{thm:eprop_converge}, which demonstrates that e-prop can converge to \(W^*\) (the solution obtained by BPTT) in a highly simplified setting, highlighting the critical role of initialization. Specifically, we present a 1D linear RNN example where e-prop matches BPTT under certain initializations but diverges under others. This toy case illustrates the sensitivity of e-prop to initialization, a key factor also observed in our empirical results. Extending such theoretical analysis to higher-dimensional RNNs would be valuable, though likely on the order of~\cite{schuessler2020interplay}, and is left for future work. Our aim is not to offer a comprehensive theoretical account of e-prop, but rather to motivate future investigation by showing that convergence behavior in even the simplest cases can depend strongly on initialization. 

\begin{prop} \label{thm:eprop_converge} (Informal)
For a 1D linear RNN trained using e-prop, the weight updates can converge to (or diverge from) the solution \( W^* \) found via an arbitrary algorithm (e.g., BPTT), depending on the initialization of the recurrent weight \( W(0) \). (The formal statement and proof are provided in Appendix~\ref{scn:proof}.)
\end{prop} 

\begin{figure*}[h!]
    \centering
    \includegraphics[trim={0cm 0cm 0cm 0cm},width=0.99\textwidth]{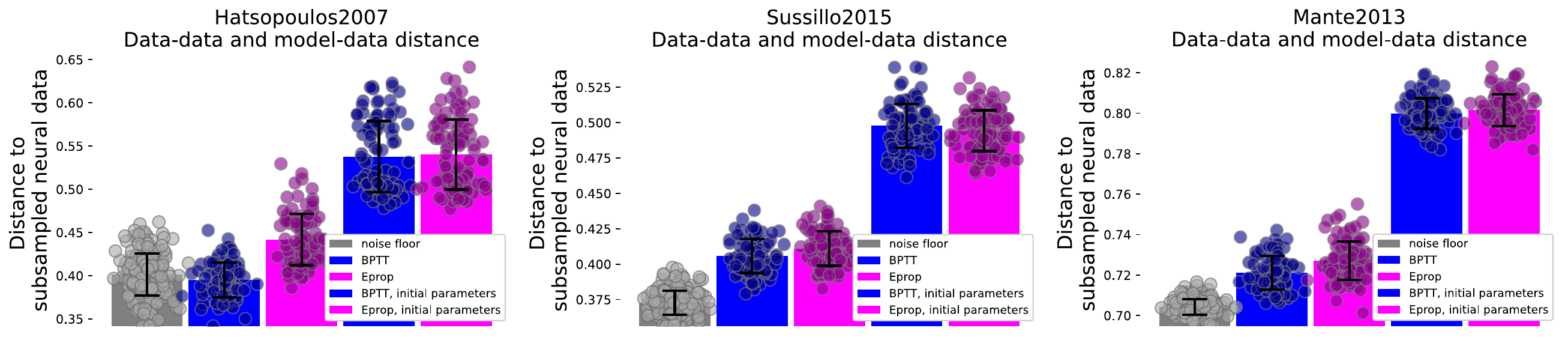}
    \caption{ Data-to-data distance (noise floor) vs model-to-data distance (BPTT and e-prop before and after training). Left: Hatsopolous 2007; middle: Sussillo 2015; right: Mante 2013. The data-splitting procedure for obtaining the baseline (i.e. noise floor) is detailed in Appendix~\ref{scn:sim_details}. We also tabularize these values in Table~\ref{tab:noise_floor}. We note that these distances are computed with fewer neurons (about half) and units than the previous plots, so the exact distance values here may differ. 
    } 
    \label{fig:noise_floor}
\end{figure*} 

\begin{table*}[ht]
\centering
\begin{tabular}{lccccc}
\toprule
\textbf{Dataset} & \textbf{Noise floor} & \textbf{BPTT (trained)} & \textbf{e-prop (trained)} & \textbf{Untrained} & \textbf{BPTT vs. floor} \\
\midrule
Hatsopoulos 2007 & 0.401~$\pm$~0.024 & 0.395~$\pm$~0.020 & 0.442~$\pm$~0.030 & 0.538~$\pm$~0.041 & p=0.311 \\
Sussillo 2015 & 0.373~$\pm$~0.009 & 0.406~$\pm$~0.012 & 0.411~$\pm$~0.012 & 0.498~$\pm$~0.015 & p=5.02e-6 \\
Mante 2013 & 0.704~$\pm$~0.004 & 0.721~$\pm$~0.008 & 0.727~$\pm$~0.010 & 0.800~$\pm$~0.008 & p=1.29e-5 \\
\bottomrule 
\end{tabular}
\caption{Neural similarity scores from Figure~\ref{fig:noise_floor}. Trained models outperform their untrained initializations and the noise floor. P-values (uncorrected t-tests) compare BPTT-trained models to the noise floor; similar trend was observed for e-prop trained model as well. Significant differences from the noise floor are found for Sussillo 2015 and Mante 2013, but not for Hatsopoulos 2007.}
\label{tab:noise_floor}
\end{table*}

It is noteworthy that if all models were equally far from the data, it might also suggest random noise. However, that is not the case, as  our models are significantly closer to the neural data after training (Figure~\ref{fig:noise_floor} and Table~\ref{tab:noise_floor}). These results are also consistent with the existing studies that reported the correlation between task performance and neural data similarity~\cite{yamins2014performance}. Additionally, what does it mean for a model to be close to the data and how do we interpret the range of alignment values? To interpret those, we need a baseline based on data-to-data similarity, which reflects how close the models are to the data relative to other data points (subsamples within the dataset). Due to limited subjects, we generated this baseline by splitting the data by neurons within the same subject, matched across timepoints and experimental conditions, though this may create an overly stringent baseline due to potential neuron dependence (details in Appendix~\ref{scn:sim_details}). We also explain why we didn't use trial-based subsampling in Appendix~\ref{scn:sim_details}. For the Hatsopoulos 2007 dataset~\cite{hatsopoulos2007encoding}, the final trained models match the neural data as closely as other neurons (Figure~\ref{fig:noise_floor}). For the Sussillo 2015 and Mante 2013 datasets, trained models approach the noise floor compared to untrained models; the remaining differences from the baseline offer insights for improving learning algorithms and architectures in future work. 

\section{Discussion}

Investigating biologically plausible learning rules is important to understanding how the brain learns. While many such rules adhere to synaptic-level constraints~\cite{richards2019deep,lillicrap2020backpropagation}, their ability to reproduce brain-like neural activity remains unclear. Here, we assess whether RNNs trained with approximate, gradient-based bio-plausible rules --- focusing on e-prop (and ModProp in the Appendix) --- can match the neural similarity of standard BPTT-trained models. Using Procrustes analysis, we show that at matched task accuracy, e-prop-trained RNNs exhibit neural similarity comparable to BPTT. Notably, architectural choices and initialization have a larger influence on model-data alignment than the choice of learning rule (Figure~\ref{fig:gain}, Appendix Figure~\ref{fig:var_cond}). Further, BPTT exhibits increased similarity to bio-plausible models at lower learning rates (Figure~\ref{fig:lrsim}) and shares post-training properties in weight eigenspectrum and dynamics, as shown via DSA (Appendix Figure~\ref{fig:W_eig}). These findings demonstrate that bio-plausible rules can, under certain conditions, achieve both strong task performance and brain-like dynamics --- motivating a broader reevaluation of how architecture and initialization shape neural similarity. 

While BPTT exhibits questionable biological plausibility, it has been widely used for brain modeling, especially in seminal works~\cite{mante2013context, sussillo2015neural, yamins2014performance, yang2020artificial}. Comparing BPTT to e-prop allows us to examine how biologically motivated gradient truncation affects neural similarity, and whether BPTT-based models, despite their widespread use, might under- or overestimate representational alignment relative to plausible alternatives. Importantly, it was not obvious a priori whether e-prop would yield representations more or less brain-like than BPTT. Our finding that e-prop can match BPTT under certain conditions is therefore nontrivial and highlights the promise of such rules. More broadly, our framework is designed to eventually evaluate models that surpass BPTT as a benchmark for neural similarity.

\textbf{Limitations and future works.} 
Extending our approach to include a broader range of learning rules, architectures, datasets, and comparison methods represents an important direction for future research. A comprehensive evaluation across these dimensions is beyond the scope of a single paper, especially in a rapidly evolving field. Our pipeline is flexible, allowing for expansion across these various facets in future investigations. Moreover, while our study is limited in the learning rules and architectures examined, it demonstrates the existence of scenarios where biologically plausible learning rules can achieve neural data similarities comparable to their deep learning counterparts. Our focus on gradient truncation-based learning rules is due to their effectiveness in task learning and versatility (see Related Works). Despite focusing primarily on e-prop (and ModProp in the Appendix), our findings suggest that it's possible for biologically plausible learning approximations to preserve synaptic-level constraints without substantially sacrificing neural data alignment. That said, existing findings do not suggest that maintaining similarity at the synaptic level necessarily translates to similarity at the neural activity level, possibly due to differences in scale and the emergent nature of neural dynamics. Additionally, while this work emphasizes learning via synaptic credit assignment, alternative frameworks such as biologically plausible in-context learning~\cite{von2023transformers} offer complementary perspectives and merit deeper investigation. 

Another limitation worth noting is that our pipeline is not designed to identify the brain’s actual learning rule. This stems from identifiability issues inherent in under-constrained systems: multiple learning rules may converge to similar terminal activity patterns when only post-learning data are available. Constraining the architecture using experimental data --- or incorporating neural data during learning --- could help narrow the solution space, and we view this as a promising direction for future work. That said, our goal is not to recover the precise rule used by the brain, but rather to test whether any existing biologically plausible rule (e.g., e-prop) can yield neural dynamics comparable to those trained with BPTT, the de facto benchmark for brain-like models. Our findings suggest that such rules can indeed achieve competitive neural similarity and task performance without sacrificing biological plausibility. 

Beyond learning rules, factors such as architecture and initialization --- highlighted in Figure~\ref{fig:gain} and Appendix Figure~\ref{fig:var_cond} --- are critical areas for future exploration. While our results show that e-prop can match BPTT in neural similarity at equal task accuracy, this does not imply the two are functionally indistinguishable. In fact, e-prop struggles on certain tasks~\cite{liu2021cell}, motivating future efforts to gather experimental data where bio-plausible rules diverge, enabling more targeted comparisons. 
This consideration partly motivated our focus on primate datasets, which typically involve more complex tasks than those used during rodent experiments.
Expanding the analysis to more open-source primate datasets and other modalities is an important next step. Relatedly, our study centers on temporal tasks with rich dynamics, aligning with our goal of investigating biologically plausible credit assignment over time. That said, extending these analyses to non-temporal domains --- such as vision, where many recent bio-plausible rules are formulated --- is a valuable direction for future work. 

As recent studies have noted, BPTT-trained models often converge to a limited subset of solutions~\cite{pagan2025individual}, and it remains unclear whether the brain uses BPTT at all~\cite{song2024inferring}; \citet{zahorodnii2025overcoming} further demonstrate that handcrafted solutions can produce more brain-like dynamics than BPTT. These observations underscore the limitations of BPTT as a benchmark and motivate future research on alternative learning frameworks and rule–architecture co-design. While in our study we intentionally fixed the architecture, initialization, and task --- following prior work (e.g.,~\cite{liu2022beyond}) --- to isolate the effect of the learning rule on neural similarity, many biologically plausible rules are designed with specific architectural constraints in mind. A more systematic investigation of learning rule and architecture interactions is an important direction for future work. 

Furthermore, we chose to focus on Procrustes distance for its ability to provide a proper metric for comparing the geometry of state representations, and its stringency in only allowing for rotations and a global stretching to align neural trajectories (although we also explored additional measures in Table~\ref{tab:additional_measures} and Appendix Figure~\ref{fig:W_eig}B). Beyond its geometric grounding, recent work~\cite{cloos2024differentiable} found that optimizing Procrustes may better preserve task-relevant features than other metrics, suggesting it can capture more meaningful neural structure. 
They also showed that Procrustes is stricter than CKA: high Procrustes was observed to imply high CKA, but not vice versa (see also~\citet{harvey2024representational}). We were also motivated to emphasize Procrustes distance because several weaknesses have been identified in other similarity measures that are, for example, biased due to high dimensionality, or may rely on low variance noise components of the data~\cite{kornblith2019similarity,Davari2022,Dujmovic2023,Elmoznino2024,cloos2024differentiable}. That said, like all scalar measures, it focuses on specific structures, and it remains uncertain whether these structures accurately capture the computational properties of interest. Relatedly, certain deep learning architectures, under sufficient conditions, may lead many models to exhibit similar internal representations~\cite{huh2024position}. Therefore, developing new measures remains a crucial and intriguing endeavor~\cite{Sexton2022, sucholutsky2023getting, Lin2023, Klabunde2023, Bowers2023, Lampinen2024}. Altogether, this vibrant area invites further work on learning rule–architecture interactions, task design, and evaluation methods. 

\section*{Acknowledgement}

We are grateful to Mante, Hatsopoulos, and Churchland for sharing their neural datasets. We also thank Nathan Cloos and Katheryn Zhou for their preprocessing code for the Mante 2013 dataset, which builds on Valerio Mante’s original scripts. This research was supported by FRQNT B3X (Y.H.L.); Pearson Fellowship (Y.H.L.). 

\section*{Impact Statement}

This research advances our understanding of biologically plausible learning models in recurrent neural networks, with no immediate ethical or societal impacts expected. Over time, the findings could influence related fields like neuroscience and deep learning, potentially affecting society based on how these disciplines evolve.

\bibliography{ref_main}

\begin{thebibliography}{112}
\providecommand{\natexlab}[1]{#1}
\providecommand{\url}[1]{\texttt{#1}}
\expandafter\ifx\csname urlstyle\endcsname\relax
  \providecommand{\doi}[1]{doi: #1}\else
  \providecommand{\doi}{doi: \begingroup \urlstyle{rm}\Url}\fi

\bibitem[Aceituno et~al.(2023)Aceituno, Farinha, Loidl, and
  Grewe]{aceituno2023learning}
Aceituno, P.~V., Farinha, M.~T., Loidl, R., and Grewe, B.~F.
\newblock Learning cortical hierarchies with temporal hebbian updates.
\newblock \emph{Frontiers in Computational Neuroscience}, 17:\penalty0 1136010,
  2023.

\bibitem[Aljadeff et~al.(2019)Aljadeff, D'amour, Field, Froemke, and
  Clopath]{aljadeff2019cortical}
Aljadeff, J., D'amour, J., Field, R.~E., Froemke, R.~C., and Clopath, C.
\newblock Cortical credit assignment by hebbian, neuromodulatory and inhibitory
  plasticity.
\newblock \emph{arXiv preprint arXiv:1911.00307}, 2019.

\bibitem[Ashwood et~al.(2020)Ashwood, Roy, Bak, and
  Pillow]{ashwood2020inferring}
Ashwood, Z., Roy, N.~A., Bak, J.~H., and Pillow, J.~W.
\newblock Inferring learning rules from animal decision-making.
\newblock \emph{Advances in Neural Information Processing Systems},
  33:\penalty0 3442--3453, 2020.

\bibitem[Barak(2017)]{barak2017recurrent}
Barak, O.
\newblock Recurrent neural networks as versatile tools of neuroscience
  research.
\newblock \emph{Current opinion in neurobiology}, 46:\penalty0 1--6, 2017.

\bibitem[Bellec et~al.(2020)Bellec, Scherr, Subramoney, Hajek, Salaj,
  Legenstein, and Maass]{bellec2020solution}
Bellec, G., Scherr, F., Subramoney, A., Hajek, E., Salaj, D., Legenstein, R.,
  and Maass, W.
\newblock A solution to the learning dilemma for recurrent networks of spiking
  neurons.
\newblock \emph{Nature communications}, 11\penalty0 (1):\penalty0 3625, 2020.

\bibitem[Bi \& Poo(1998)Bi and Poo]{bi1998synaptic}
Bi, G.-q. and Poo, M.-m.
\newblock Synaptic modifications in cultured hippocampal neurons: dependence on
  spike timing, synaptic strength, and postsynaptic cell type.
\newblock \emph{Journal of neuroscience}, 18\penalty0 (24):\penalty0
  10464--10472, 1998.

\bibitem[Bordelon \& Pehlevan(2022)Bordelon and
  Pehlevan]{bordelon2022influence}
Bordelon, B. and Pehlevan, C.
\newblock The influence of learning rule on representation dynamics in wide
  neural networks.
\newblock \emph{arXiv preprint arXiv:2210.02157}, 2022.

\bibitem[Bowers et~al.(2023)Bowers, Malhotra, Dujmović, Llera~Montero,
  Tsvetkov, Biscione, Puebla, Adolfi, Hummel, Heaton, and et~al.]{Bowers2023}
Bowers, J.~S., Malhotra, G., Dujmović, M., Llera~Montero, M., Tsvetkov, C.,
  Biscione, V., Puebla, G., Adolfi, F., Hummel, J.~E., Heaton, R.~F., and
  et~al.
\newblock Deep problems with neural network models of human vision.
\newblock \emph{Behavioral and Brain Sciences}, 46:\penalty0 e385, 2023.

\bibitem[Braun et~al.(2022)Braun, Domin{\'e}, Fitzgerald, and
  Saxe]{braun2022exact}
Braun, L., Domin{\'e}, C., Fitzgerald, J., and Saxe, A.
\newblock Exact learning dynamics of deep linear networks with prior knowledge.
\newblock \emph{Advances in Neural Information Processing Systems},
  35:\penalty0 6615--6629, 2022.

\bibitem[Bredenberg et~al.(2023)Bredenberg, Williams, Savin, Richards, and
  Lajoie]{bredenberg2023formalizing}
Bredenberg, C., Williams, E., Savin, C., Richards, B., and Lajoie, G.
\newblock Formalizing locality for normative synaptic plasticity models.
\newblock \emph{Advances in Neural Information Processing Systems},
  36:\penalty0 5653--5684, 2023.

\bibitem[Chaudhuri et~al.(2019)Chaudhuri, Ger{\c{c}}ek, Pandey, Peyrache, and
  Fiete]{chaudhuri2019intrinsic}
Chaudhuri, R., Ger{\c{c}}ek, B., Pandey, B., Peyrache, A., and Fiete, I.
\newblock The intrinsic attractor manifold and population dynamics of a
  canonical cognitive circuit across waking and sleep.
\newblock \emph{Nature neuroscience}, 22\penalty0 (9):\penalty0 1512--1520,
  2019.

\bibitem[Chizat et~al.(2019)Chizat, Oyallon, and Bach]{chizat2019lazy}
Chizat, L., Oyallon, E., and Bach, F.
\newblock On lazy training in differentiable programming.
\newblock \emph{Advances in Neural Information Processing Systems}, 32, 2019.

\bibitem[Chung \& Abbott(2021)Chung and Abbott]{chung2021neural}
Chung, S. and Abbott, L.~F.
\newblock Neural population geometry: An approach for understanding biological
  and artificial neural networks.
\newblock \emph{Current opinion in neurobiology}, 70:\penalty0 137--144, 2021.

\bibitem[Cloos et~al.(2022)Cloos, Li, Yang, and Cueva]{cloos2022scaling}
Cloos, N., Li, M., Yang, G.~R., and Cueva, C.~J.
\newblock Scaling up the evaluation of recurrent neural network models for
  cognitive neuroscience.
\newblock \emph{Cognitive Computational Neuroscience}, 2022.

\bibitem[Cloos et~al.(2024)Cloos, Li, Siegel, Brincat, Miller, Yang, and
  Cueva]{cloos2024differentiable}
Cloos, N., Li, M., Siegel, M., Brincat, S.~L., Miller, E.~K., Yang, G.~R., and
  Cueva, C.~J.
\newblock Differentiable optimization of similarity scores between models and
  brains.
\newblock \emph{arXiv preprint arXiv:2407.07059}, 2024.

\bibitem[Codol et~al.(2024)Codol, Krishna, Lajoie, and Perich]{codol2024brain}
Codol, O., Krishna, N.~H., Lajoie, G., and Perich, M.~G.
\newblock Brain-like neural dynamics for behavioral control develop through
  reinforcement learning.
\newblock \emph{bioRxiv}, 2024.

\bibitem[Confavreux et~al.(2024)Confavreux, Ramesh, Goncalves, Macke, and
  Vogels]{confavreux2024meta}
Confavreux, B., Ramesh, P., Goncalves, P.~J., Macke, J.~H., and Vogels, T.
\newblock Meta-learning families of plasticity rules in recurrent spiking
  networks using simulation-based inference.
\newblock \emph{Advances in Neural Information Processing Systems}, 36, 2024.

\bibitem[Cunningham et~al.(2009)Cunningham, Gilja, Ryu, and
  Shenoy]{cunningham2009methods}
Cunningham, J.~P., Gilja, V., Ryu, S.~I., and Shenoy, K.~V.
\newblock Methods for estimating neural firing rates, and their application to
  brain--machine interfaces.
\newblock \emph{Neural Networks}, 22\penalty0 (9):\penalty0 1235--1246, 2009.

\bibitem[Davari et~al.(2022)Davari, Horoi, Natik, Lajoie, Wolf, and
  Belilovsky]{Davari2022}
Davari, M., Horoi, S., Natik, A., Lajoie, G., Wolf, G., and Belilovsky, E.
\newblock Reliability of cka as a similarity measure in deep learning.
\newblock \emph{arXiv preprint arXiv:2210.16156}, 2022.

\bibitem[Degenhart et~al.(2020)Degenhart, Bishop, Oby, Tyler-Kabara, Chase,
  Batista, and Yu]{degenhart2020stabilization}
Degenhart, A.~D., Bishop, W.~E., Oby, E.~R., Tyler-Kabara, E.~C., Chase, S.~M.,
  Batista, A.~P., and Yu, B.~M.
\newblock Stabilization of a brain--computer interface via the alignment of
  low-dimensional spaces of neural activity.
\newblock \emph{Nature biomedical engineering}, 4\penalty0 (7):\penalty0
  672--685, 2020.

\bibitem[DePasquale et~al.(2018)DePasquale, Cueva, Rajan, Escola, and
  Abbott]{depasquale2018full}
DePasquale, B., Cueva, C.~J., Rajan, K., Escola, G.~S., and Abbott, L.
\newblock full-force: A target-based method for training recurrent networks.
\newblock \emph{PloS one}, 13\penalty0 (2):\penalty0 e0191527, 2018.

\bibitem[Ding et~al.(2021)Ding, Denain, and Steinhardt]{ding2021grounding}
Ding, F., Denain, J.-S., and Steinhardt, J.
\newblock Grounding representation similarity through statistical testing.
\newblock \emph{Advances in Neural Information Processing Systems},
  34:\penalty0 1556--1568, 2021.

\bibitem[Dujmovi{\'c} et~al.(2023)Dujmovi{\'c}, Bowers, Adolfi, and
  Malhotra]{Dujmovic2023}
Dujmovi{\'c}, M., Bowers, J.~S., Adolfi, F., and Malhotra, G.
\newblock Obstacles to inferring mechanistic similarity using representational
  similarity analysis.
\newblock \emph{bioRxiv}, 2023.

\bibitem[Duong et~al.(2022)Duong, Zhou, Nassar, Berman, Olieslagers, and
  Williams]{duong2022representational}
Duong, L.~R., Zhou, J., Nassar, J., Berman, J., Olieslagers, J., and Williams,
  A.~H.
\newblock Representational dissimilarity metric spaces for stochastic neural
  networks.
\newblock \emph{arXiv preprint arXiv:2211.11665}, 2022.

\bibitem[Elmoznino \& Bonner(2024)Elmoznino and Bonner]{Elmoznino2024}
Elmoznino, E. and Bonner, M.~F.
\newblock High-performing neural network models of visual cortex benefit from
  high latent dimensionality.
\newblock \emph{PLOS Computational Biology}, 20\penalty0 (1):\penalty0 1--23,
  01 2024.

\bibitem[Eyono et~al.(2022)Eyono, Boven, Ghosh, Pemberton, Scherr, Clopath,
  Costa, Maass, Richards, Savin, et~al.]{eyono2022current}
Eyono, R.~H., Boven, E., Ghosh, A., Pemberton, J., Scherr, F., Clopath, C.,
  Costa, R.~P., Maass, W., Richards, B.~A., Savin, C., et~al.
\newblock Current state and future directions for learning in biological
  recurrent neural networks: A perspective piece.
\newblock \emph{Neurons, Behavior, Data analysis, and Theory}, 1, 2022.

\bibitem[Fetz(1992)]{Fetz1992}
Fetz, E.
\newblock Are movement parameters recognizably coded in the activity of single
  neurons?
\newblock \emph{Behavioral and Brain Sciences}, 15\penalty0 (4):\penalty0
  679--690, 1992.

\bibitem[Flesch et~al.(2023)Flesch, Saxe, and Summerfield]{flesch2023continual}
Flesch, T., Saxe, A., and Summerfield, C.
\newblock Continual task learning in natural and artificial agents.
\newblock \emph{Trends in Neurosciences}, 46\penalty0 (3):\penalty0 199--210,
  2023.

\bibitem[Gerstner et~al.(2018)Gerstner, Lehmann, Liakoni, Corneil, and
  Brea]{Gerstner2018}
Gerstner, W., Lehmann, M., Liakoni, V., Corneil, D., and Brea, J.
\newblock {Eligibility Traces and Plasticity on Behavioral Time Scales:
  Experimental Support of NeoHebbian Three-Factor Learning Rules}.
\newblock \emph{Frontiers in Neural Circuits}, 12:\penalty0 53, jul 2018.

\bibitem[Ghosh et~al.(2023)Ghosh, Liu, Lajoie, Kording, and
  Richards]{ghosh2023gradient}
Ghosh, A., Liu, Y.~H., Lajoie, G., Kording, K., and Richards, B.~A.
\newblock How gradient estimator variance and bias impact learning in neural
  networks.
\newblock In \emph{The Eleventh International Conference on Learning
  Representations}, 2023.

\bibitem[Gilra \& Gerstner(2017)Gilra and Gerstner]{gilra2017predicting}
Gilra, A. and Gerstner, W.
\newblock Predicting non-linear dynamics by stable local learning in a
  recurrent spiking neural network.
\newblock \emph{Elife}, 6:\penalty0 e28295, 2017.

\bibitem[Greedy et~al.(2022)Greedy, Zhu, Pemberton, Mellor, and
  Ponte~Costa]{greedy2022single}
Greedy, W., Zhu, H.~W., Pemberton, J., Mellor, J., and Ponte~Costa, R.
\newblock Single-phase deep learning in cortico-cortical networks.
\newblock \emph{Advances in Neural Information Processing Systems},
  35:\penalty0 24213--24225, 2022.

\bibitem[Harvey et~al.(2023)Harvey, Larsen, and Williams]{Harvey2023}
Harvey, S.~E., Larsen, B.~W., and Williams, A.~H.
\newblock Duality of bures and shape distances with implications for comparing
  neural representations.
\newblock In \emph{UniReps: the First Workshop on Unifying Representations in
  Neural Models}, 2023.

\bibitem[Harvey et~al.(2024)Harvey, Lipshutz, and
  Williams]{harvey2024representational}
Harvey, S.~E., Lipshutz, D., and Williams, A.~H.
\newblock What representational similarity measures imply about decodable
  information.
\newblock \emph{arXiv preprint arXiv:2411.08197}, 2024.

\bibitem[Hatsopoulos et~al.(2007)Hatsopoulos, Xu, and
  Amit]{hatsopoulos2007encoding}
Hatsopoulos, N.~G., Xu, Q., and Amit, Y.
\newblock Encoding of movement fragments in the motor cortex.
\newblock \emph{Journal of Neuroscience}, 27\penalty0 (19):\penalty0
  5105--5114, 2007.

\bibitem[Hinton(2022)]{hinton2022forward}
Hinton, G.
\newblock The forward-forward algorithm: Some preliminary investigations.
\newblock \emph{arXiv preprint arXiv:2212.13345}, 2022.

\bibitem[Huh et~al.(2024)Huh, Cheung, Wang, and Isola]{huh2024position}
Huh, M., Cheung, B., Wang, T., and Isola, P.
\newblock Position: The platonic representation hypothesis.
\newblock In \emph{Forty-first International Conference on Machine Learning},
  2024.

\bibitem[Kaleb et~al.(2024)Kaleb, Feulner, Gallego, and
  Clopath]{kaleb2024feedback}
Kaleb, K., Feulner, B., Gallego, J., and Clopath, C.
\newblock Feedback control guides credit assignment in recurrent neural
  networks.
\newblock \emph{Advances in Neural Information Processing Systems},
  37:\penalty0 5122--5144, 2024.

\bibitem[Kay et~al.(2024)Kay, Prince, Gebhart, Tuckute, Zhou, Naselaris, and
  Schutt]{kay2024disentangling}
Kay, K., Prince, J.~S., Gebhart, T., Tuckute, G., Zhou, J., Naselaris, T., and
  Schutt, H.
\newblock Disentangling signal and noise in neural responses through generative
  modeling.
\newblock \emph{bioRxiv}, 2024.

\bibitem[Kepple et~al.(2021)Kepple, Engelken, and Rajan]{kepple2021curriculum}
Kepple, D.~R., Engelken, R., and Rajan, K.
\newblock Curriculum learning as a tool to uncover learning principles in the
  brain.
\newblock In \emph{International Conference on Learning Representations}, 2021.

\bibitem[Khosla \& Williams(2023)Khosla and Williams]{khosla2023soft}
Khosla, M. and Williams, A.~H.
\newblock Soft matching distance: A metric on neural representations that
  captures single-neuron tuning.
\newblock \emph{arXiv preprint arXiv:2311.09466}, 2023.

\bibitem[Klabunde et~al.(2023)Klabunde, Schumacher, Strohmaier, and
  Lemmerich]{Klabunde2023}
Klabunde, M., Schumacher, T., Strohmaier, M., and Lemmerich, F.
\newblock Similarity of neural network models: A survey of functional and
  representational measures.
\newblock \emph{arXiv preprint arXiv:2305.06329}, 2023.

\bibitem[Kornblith et~al.(2019)Kornblith, Norouzi, Lee, and
  Hinton]{kornblith2019similarity}
Kornblith, S., Norouzi, M., Lee, H., and Hinton, G.
\newblock Similarity of neural network representations revisited.
\newblock In \emph{International conference on machine learning}, pp.\
  3519--3529. PMLR, 2019.

\bibitem[Kriegeskorte et~al.(2008)Kriegeskorte, Mur, and
  Bandettini]{kriegeskorte2008representational}
Kriegeskorte, N., Mur, M., and Bandettini, P.~A.
\newblock Representational similarity analysis-connecting the branches of
  systems neuroscience.
\newblock \emph{Frontiers in systems neuroscience}, 2:\penalty0 249, 2008.

\bibitem[Laborieux \& Zenke(2022)Laborieux and Zenke]{laborieux2022holomorphic}
Laborieux, A. and Zenke, F.
\newblock Holomorphic equilibrium propagation computes exact gradients through
  finite size oscillations.
\newblock \emph{arXiv preprint arXiv:2209.00530}, 2022.

\bibitem[Lampinen et~al.(2024)Lampinen, Chan, and Hermann]{Lampinen2024}
Lampinen, A.~K., Chan, S. C.~Y., and Hermann, K.
\newblock Learned feature representations are biased by complexity, learning
  order, position, and more.
\newblock \emph{arXiv preprint arXiv:2405.05847}, 2024.

\bibitem[Langdon \& Engel(2022)Langdon and Engel]{langdon2022latent}
Langdon, C. and Engel, T.~A.
\newblock Latent circuit inference from heterogeneous neural responses during
  cognitive tasks.
\newblock \emph{bioRxiv}, 2022.

\bibitem[Lillicrap et~al.(2016)Lillicrap, Cownden, Tweed, and
  Akerman]{lillicrap2016random}
Lillicrap, T.~P., Cownden, D., Tweed, D.~B., and Akerman, C.~J.
\newblock Random synaptic feedback weights support error backpropagation for
  deep learning.
\newblock \emph{Nature communications}, 7\penalty0 (1):\penalty0 1--10, 2016.

\bibitem[Lillicrap et~al.(2020)Lillicrap, Santoro, Marris, Akerman, and
  Hinton]{lillicrap2020backpropagation}
Lillicrap, T.~P., Santoro, A., Marris, L., Akerman, C.~J., and Hinton, G.
\newblock Backpropagation and the brain.
\newblock \emph{Nature Reviews Neuroscience}, 21\penalty0 (6):\penalty0
  335--346, 2020.

\bibitem[Lim et~al.(2015)Lim, McKee, Woloszyn, Amit, Freedman, Sheinberg, and
  Brunel]{lim2015inferring}
Lim, S., McKee, J.~L., Woloszyn, L., Amit, Y., Freedman, D.~J., Sheinberg,
  D.~L., and Brunel, N.
\newblock Inferring learning rules from distributions of firing rates in
  cortical neurons.
\newblock \emph{Nature neuroscience}, 18\penalty0 (12):\penalty0 1804--1810,
  2015.

\bibitem[Lin \& Kriegeskorte(2023{\natexlab{a}})Lin and Kriegeskorte]{Lin2023}
Lin, B. and Kriegeskorte, N.
\newblock The topology and geometry of neural representations.
\newblock \emph{arXiv preprint arXiv:2309.11028}, 2023{\natexlab{a}}.

\bibitem[Lin \& Kriegeskorte(2023{\natexlab{b}})Lin and
  Kriegeskorte]{lin2023topology}
Lin, B. and Kriegeskorte, N.
\newblock The topology and geometry of neural representations.
\newblock \emph{arXiv preprint arXiv:2309.11028}, 2023{\natexlab{b}}.

\bibitem[Liu et~al.(2021)Liu, Smith, Mihalas, Shea-Brown, and
  S{\"u}mb{\"u}l]{liu2021cell}
Liu, Y.~H., Smith, S., Mihalas, S., Shea-Brown, E., and S{\"u}mb{\"u}l, U.
\newblock Cell-type--specific neuromodulation guides synaptic credit assignment
  in a spiking neural network.
\newblock \emph{Proceedings of the National Academy of Sciences}, 118\penalty0
  (51):\penalty0 e2111821118, 2021.

\bibitem[Liu et~al.(2022{\natexlab{a}})Liu, Ghosh, Richards, Shea-Brown, and
  Lajoie]{liu2022beyond}
Liu, Y.~H., Ghosh, A., Richards, B.~A., Shea-Brown, E., and Lajoie, G.
\newblock Beyond accuracy: generalization properties of bio-plausible temporal
  credit assignment rules.
\newblock \emph{arXiv preprint arXiv:2206.00823}, 2022{\natexlab{a}}.

\bibitem[Liu et~al.(2022{\natexlab{b}})Liu, Smith, Mihalas, Shea-Brown, and
  S{\"u}mb{\"u}l]{liu2022biologically}
Liu, Y.~H., Smith, S., Mihalas, S., Shea-Brown, E., and S{\"u}mb{\"u}l, U.
\newblock Biologically-plausible backpropagation through arbitrary timespans
  via local neuromodulators.
\newblock \emph{arXiv preprint arXiv:2206.01338}, 2022{\natexlab{b}}.

\bibitem[Liu et~al.(2023)Liu, Baratin, Cornford, Mihalas, Shea-Brown, and
  Lajoie]{liu2023connectivity}
Liu, Y.~H., Baratin, A., Cornford, J., Mihalas, S., Shea-Brown, E., and Lajoie,
  G.
\newblock How connectivity structure shapes rich and lazy learning in neural
  circuits.
\newblock \emph{arXiv preprint arXiv:2310.08513}, 2023.

\bibitem[Magee \& Grienberger(2020)Magee and Grienberger]{Magee2020}
Magee, J.~C. and Grienberger, C.
\newblock {Synaptic Plasticity Forms and Functions}.
\newblock \emph{Annual Review of Neuroscience}, 43\penalty0 (1):\penalty0
  95--117, jul 2020.

\bibitem[Maheswaranathan et~al.(2019)Maheswaranathan, Williams, Golub, Ganguli,
  and Sussillo]{maheswaranathan2019universality}
Maheswaranathan, N., Williams, A., Golub, M., Ganguli, S., and Sussillo, D.
\newblock Universality and individuality in neural dynamics across large
  populations of recurrent networks.
\newblock \emph{Advances in neural information processing systems}, 32, 2019.

\bibitem[Mante et~al.(2013)Mante, Sussillo, Shenoy, and
  Newsome]{mante2013context}
Mante, V., Sussillo, D., Shenoy, K.~V., and Newsome, W.~T.
\newblock Context-dependent computation by recurrent dynamics in prefrontal
  cortex.
\newblock \emph{nature}, 503\penalty0 (7474):\penalty0 78--84, 2013.

\bibitem[Marschall et~al.(2019)Marschall, Cho, and
  Savin]{marschall2019evaluating}
Marschall, O., Cho, K., and Savin, C.
\newblock Evaluating biological plausibility of learning algorithms the lazy
  way.
\newblock In \emph{Real Neurons $\{$$\backslash$\&$\}$ Hidden Units: Future
  directions at the intersection of neuroscience and artificial intelligence@
  NeurIPS 2019}, 2019.

\bibitem[Marschall et~al.(2020)Marschall, Cho, and Savin]{marschall2020unified}
Marschall, O., Cho, K., and Savin, C.
\newblock A unified framework of online learning algorithms for training
  recurrent neural networks.
\newblock \emph{The Journal of Machine Learning Research}, 21\penalty0
  (1):\penalty0 5320--5353, 2020.

\bibitem[Menick et~al.(2020)Menick, Elsen, Evci, Osindero, Simonyan, and
  Graves]{Menick2020}
Menick, J., Elsen, E., Evci, U., Osindero, S., Simonyan, K., and Graves, A.
\newblock A practical sparse approximation for real time recurrent learning.
\newblock \emph{arXiv preprint arXiv:2006.07232}, 2020.

\bibitem[Meulemans et~al.(2022)Meulemans, Zucchet, Kobayashi, Von~Oswald, and
  Sacramento]{meulemans2022least}
Meulemans, A., Zucchet, N., Kobayashi, S., Von~Oswald, J., and Sacramento, J.
\newblock The least-control principle for local learning at equilibrium.
\newblock \emph{Advances in Neural Information Processing Systems},
  35:\penalty0 33603--33617, 2022.

\bibitem[Miconi(2017)]{miconi2017biologically}
Miconi, T.
\newblock Biologically plausible learning in recurrent neural networks
  reproduces neural dynamics observed during cognitive tasks.
\newblock \emph{Elife}, 6:\penalty0 e20899, 2017.

\bibitem[Molano-Mazon et~al.(2022)Molano-Mazon, Barbosa, Pastor-Ciurana,
  Fradera, Zhang, Forest, del Pozo~Lerida, Ji-An, Cueva, de~la Rocha,
  et~al.]{molano2022neurogym}
Molano-Mazon, M., Barbosa, J., Pastor-Ciurana, J., Fradera, M., Zhang, R.-Y.,
  Forest, J., del Pozo~Lerida, J., Ji-An, L., Cueva, C.~J., de~la Rocha, J.,
  et~al.
\newblock Neurogym: An open resource for developing and sharing neuroscience
  tasks.
\newblock \emph{PsyArXiv}, 2022.

\bibitem[Moody et~al.(1998)Moody, Wise, di~Pellegrino, and Zipser]{Moody1998}
Moody, S.~L., Wise, S.~P., di~Pellegrino, G., and Zipser, D.
\newblock A model that accounts for activity in primate frontal cortex during a
  delayed matching-to-sample task.
\newblock \emph{Journal of Neuroscience}, 18\penalty0 (1):\penalty0 399--410,
  1998.

\bibitem[Murray(2019)]{murray2019local}
Murray, J.~M.
\newblock Local online learning in recurrent networks with random feedback.
\newblock \emph{Elife}, 8:\penalty0 e43299, 2019.

\bibitem[Nayebi et~al.(2020)Nayebi, Srivastava, Ganguli, and
  Yamins]{nayebi2020identifying}
Nayebi, A., Srivastava, S., Ganguli, S., and Yamins, D.~L.
\newblock Identifying learning rules from neural network observables.
\newblock \emph{Advances in Neural Information Processing Systems},
  33:\penalty0 2639--2650, 2020.

\bibitem[Nejatbakhsh et~al.(2024)Nejatbakhsh, Geadah, Williams, and
  Lipshutz]{nejatbakhsh2024comparing}
Nejatbakhsh, A., Geadah, V., Williams, A.~H., and Lipshutz, D.
\newblock Comparing noisy neural population dynamics using optimal transport
  distances.
\newblock \emph{arXiv preprint arXiv:2412.14421}, 2024.

\bibitem[Ostrow et~al.(2024)Ostrow, Eisen, Kozachkov, and
  Fiete]{ostrow2024beyond}
Ostrow, M., Eisen, A., Kozachkov, L., and Fiete, I.
\newblock Beyond geometry: Comparing the temporal structure of computation in
  neural circuits with dynamical similarity analysis.
\newblock \emph{Advances in Neural Information Processing Systems}, 36, 2024.

\bibitem[Paccolat et~al.(2021)Paccolat, Petrini, Geiger, Tyloo, and
  Wyart]{paccolat2021geometric}
Paccolat, J., Petrini, L., Geiger, M., Tyloo, K., and Wyart, M.
\newblock Geometric compression of invariant manifolds in neural networks.
\newblock \emph{Journal of Statistical Mechanics: Theory and Experiment},
  2021\penalty0 (4):\penalty0 044001, 2021.

\bibitem[Pagan et~al.(2025)Pagan, Tang, Aoi, Pillow, Mante, Sussillo, and
  Brody]{pagan2025individual}
Pagan, M., Tang, V.~D., Aoi, M.~C., Pillow, J.~W., Mante, V., Sussillo, D., and
  Brody, C.~D.
\newblock Individual variability of neural computations underlying flexible
  decisions.
\newblock \emph{Nature}, 639\penalty0 (8054):\penalty0 421--429, 2025.

\bibitem[Pandarinath et~al.(2018)Pandarinath, O’Shea, Collins, Jozefowicz,
  Stavisky, Kao, Trautmann, Kaufman, Ryu, Hochberg,
  et~al.]{pandarinath2018inferring}
Pandarinath, C., O’Shea, D.~J., Collins, J., Jozefowicz, R., Stavisky, S.~D.,
  Kao, J.~C., Trautmann, E.~M., Kaufman, M.~T., Ryu, S.~I., Hochberg, L.~R.,
  et~al.
\newblock Inferring single-trial neural population dynamics using sequential
  auto-encoders.
\newblock \emph{Nature methods}, 15\penalty0 (10):\penalty0 805--815, 2018.

\bibitem[Paszke et~al.(2019)Paszke, Gross, Massa, Lerer, Bradbury, Chanan,
  Killeen, Lin, Gimelshein, Antiga, et~al.]{paszke2019pytorch}
Paszke, A., Gross, S., Massa, F., Lerer, A., Bradbury, J., Chanan, G., Killeen,
  T., Lin, Z., Gimelshein, N., Antiga, L., et~al.
\newblock Pytorch: An imperative style, high-performance deep learning library.
\newblock \emph{Advances in neural information processing systems}, 32, 2019.

\bibitem[Payeur et~al.(2021)Payeur, Guerguiev, Zenke, Richards, and
  Naud]{payeur2020burst}
Payeur, A., Guerguiev, J., Zenke, F., Richards, B.~A., and Naud, R.
\newblock Burst-dependent synaptic plasticity can coordinate learning in
  hierarchical circuits.
\newblock \emph{Nature neuroscience}, pp.\  1--10, 2021.

\bibitem[Perich et~al.(2021)Perich, Arlt, Soares, Young, Mosher, Minxha,
  Carter, Rutishauser, Rudebeck, Harvey, et~al.]{perich2021inferring}
Perich, M.~G., Arlt, C., Soares, S., Young, M.~E., Mosher, C.~P., Minxha, J.,
  Carter, E., Rutishauser, U., Rudebeck, P.~H., Harvey, C.~D., et~al.
\newblock Inferring brain-wide interactions using data-constrained recurrent
  neural network models.
\newblock \emph{bioRxiv}, pp.\  2020--12, 2021.

\bibitem[Portes et~al.(2022)Portes, Schmid, and
  Murray]{portes2022distinguishing}
Portes, J., Schmid, C., and Murray, J.~M.
\newblock Distinguishing learning rules with brain machine interfaces.
\newblock \emph{Advances in neural information processing systems},
  35:\penalty0 25937--25950, 2022.

\bibitem[Pospisil et~al.(2023)Pospisil, Larsen, Harvey, and
  Williams]{pospisil2023estimating}
Pospisil, D.~A., Larsen, B.~W., Harvey, S.~E., and Williams, A.~H.
\newblock Estimating shape distances on neural representations with limited
  samples.
\newblock \emph{arXiv preprint arXiv:2310.05742}, 2023.

\bibitem[Raghu et~al.(2017)Raghu, Gilmer, Yosinski, and
  Sohl-Dickstein]{raghu2017svcca}
Raghu, M., Gilmer, J., Yosinski, J., and Sohl-Dickstein, J.
\newblock Svcca: Singular vector canonical correlation analysis for deep
  learning dynamics and interpretability.
\newblock \emph{Advances in neural information processing systems}, 30, 2017.

\bibitem[Remy \& Spruston(2007)Remy and Spruston]{remy2007dendritic}
Remy, S. and Spruston, N.
\newblock Dendritic spikes induce single-burst long-term potentiation.
\newblock \emph{Proceedings of the National Academy of Sciences}, 104\penalty0
  (43):\penalty0 17192--17197, 2007.

\bibitem[Richards et~al.(2019)Richards, Lillicrap, Beaudoin, Bengio, Bogacz,
  Christensen, Clopath, Costa, de~Berker, Ganguli, et~al.]{richards2019deep}
Richards, B.~A., Lillicrap, T.~P., Beaudoin, P., Bengio, Y., Bogacz, R.,
  Christensen, A., Clopath, C., Costa, R.~P., de~Berker, A., Ganguli, S.,
  et~al.
\newblock A deep learning framework for neuroscience.
\newblock \emph{Nature neuroscience}, 22\penalty0 (11):\penalty0 1761--1770,
  2019.

\bibitem[Roelfsema \& Holtmaat(2018)Roelfsema and
  Holtmaat]{roelfsema2018control}
Roelfsema, P.~R. and Holtmaat, A.
\newblock Control of synaptic plasticity in deep cortical networks.
\newblock \emph{Nature Reviews Neuroscience}, 19\penalty0 (3):\penalty0
  166--180, 2018.

\bibitem[Sacramento et~al.(2018)Sacramento, Costa, Bengio, and
  Senn]{sacramento2018dendritic}
Sacramento, J., Costa, R.~P., Bengio, Y., and Senn, W.
\newblock Dendritic cortical microcircuits approximate the backpropagation
  algorithm.
\newblock \emph{arXiv preprint arXiv:1810.11393}, 2018.

\bibitem[Salimans et~al.(2017)Salimans, Ho, Chen, Sidor, and
  Sutskever]{salimans2017evolution}
Salimans, T., Ho, J., Chen, X., Sidor, S., and Sutskever, I.
\newblock Evolution strategies as a scalable alternative to reinforcement
  learning.
\newblock \emph{arXiv preprint arXiv:1703.03864}, 2017.

\bibitem[Scellier \& Bengio(2017)Scellier and Bengio]{scellier2017equilibrium}
Scellier, B. and Bengio, Y.
\newblock Equilibrium propagation: Bridging the gap between energy-based models
  and backpropagation.
\newblock \emph{Frontiers in computational neuroscience}, 11:\penalty0 24,
  2017.

\bibitem[Schrimpf et~al.(2018)Schrimpf, Kubilius, Hong, Majaj, Rajalingham,
  Issa, Kar, Bashivan, Prescott-Roy, Geiger, et~al.]{schrimpf2018brain}
Schrimpf, M., Kubilius, J., Hong, H., Majaj, N.~J., Rajalingham, R., Issa,
  E.~B., Kar, K., Bashivan, P., Prescott-Roy, J., Geiger, F., et~al.
\newblock Brain-score: Which artificial neural network for object recognition
  is most brain-like?
\newblock \emph{BioRxiv}, 2018.

\bibitem[Schuessler et~al.(2020)Schuessler, Mastrogiuseppe, Dubreuil, Ostojic,
  and Barak]{schuessler2020interplay}
Schuessler, F., Mastrogiuseppe, F., Dubreuil, A., Ostojic, S., and Barak, O.
\newblock The interplay between randomness and structure during learning in
  rnns.
\newblock \emph{Advances in neural information processing systems},
  33:\penalty0 13352--13362, 2020.

\bibitem[Schuessler et~al.(2023)Schuessler, Mastrogiuseppe, Ostojic, and
  Barak]{schuessler2023aligned}
Schuessler, F., Mastrogiuseppe, F., Ostojic, S., and Barak, O.
\newblock Aligned and oblique dynamics in recurrent neural networks.
\newblock \emph{arXiv preprint arXiv:2307.07654}, 2023.

\bibitem[Sexton \& Love(2022)Sexton and Love]{Sexton2022}
Sexton, N.~J. and Love, B.~C.
\newblock Reassessing hierarchical correspondences between brain and deep
  networks through direct interface.
\newblock \emph{Science Advances}, 8\penalty0 (28), 2022.

\bibitem[Shahbazi et~al.(2021)Shahbazi, Shirali, Aghajan, and
  Nili]{shahbazi2021using}
Shahbazi, M., Shirali, A., Aghajan, H., and Nili, H.
\newblock Using distance on the riemannian manifold to compare representations
  in brain and in models.
\newblock \emph{NeuroImage}, 239:\penalty0 118271, 2021.

\bibitem[Smith et~al.(2020)Smith, Hawrylycz, Rossier, and
  S{\"{u}}mb{\"{u}}l]{Smith2020}
Smith, S.~J., Hawrylycz, M., Rossier, J., and S{\"{u}}mb{\"{u}}l, U.
\newblock {New light on cortical neuropeptides and synaptic network
  plasticity}.
\newblock \emph{Current Opinion in Neurobiology}, 63:\penalty0 176--188, aug
  2020.

\bibitem[Song et~al.(2016)Song, Yang, and Wang]{song2016training}
Song, H.~F., Yang, G.~R., and Wang, X.-J.
\newblock Training excitatory-inhibitory recurrent neural networks for
  cognitive tasks: a simple and flexible framework.
\newblock \emph{PLOS Computational Biology}, 12\penalty0 (2):\penalty0 1--30,
  02 2016.

\bibitem[Song et~al.(2024)Song, Millidge, Salvatori, Lukasiewicz, Xu, and
  Bogacz]{song2024inferring}
Song, Y., Millidge, B., Salvatori, T., Lukasiewicz, T., Xu, Z., and Bogacz, R.
\newblock Inferring neural activity before plasticity as a foundation for
  learning beyond backpropagation.
\newblock \emph{Nature Neuroscience}, 27\penalty0 (2):\penalty0 348--358, 2024.

\bibitem[Sucholutsky et~al.(2023)Sucholutsky, Muttenthaler, Weller, Peng, Bobu,
  Kim, Love, Grant, Achterberg, Tenenbaum, et~al.]{sucholutsky2023getting}
Sucholutsky, I., Muttenthaler, L., Weller, A., Peng, A., Bobu, A., Kim, B.,
  Love, B.~C., Grant, E., Achterberg, J., Tenenbaum, J.~B., et~al.
\newblock Getting aligned on representational alignment.
\newblock \emph{arXiv preprint arXiv:2310.13018}, 2023.

\bibitem[Sussillo \& Abbott(2009)Sussillo and Abbott]{sussillo2009generating}
Sussillo, D. and Abbott, L.~F.
\newblock Generating coherent patterns of activity from chaotic neural
  networks.
\newblock \emph{Neuron}, 63\penalty0 (4):\penalty0 544--557, 2009.

\bibitem[Sussillo et~al.(2015)Sussillo, Churchland, Kaufman, and
  Shenoy]{sussillo2015neural}
Sussillo, D., Churchland, M.~M., Kaufman, M.~T., and Shenoy, K.~V.
\newblock A neural network that finds a naturalistic solution for the
  production of muscle activity.
\newblock \emph{Nature neuroscience}, 18\penalty0 (7):\penalty0 1025--1033,
  2015.

\bibitem[Turner et~al.(2021)Turner, Dabholkar, and Barak]{turner2021charting}
Turner, E., Dabholkar, K.~V., and Barak, O.
\newblock Charting and navigating the space of solutions for recurrent neural
  networks.
\newblock \emph{Advances in Neural Information Processing Systems},
  34:\penalty0 25320--25333, 2021.

\bibitem[Valente et~al.(2021)Valente, Ostojic, and Pillow]{valente2021probing}
Valente, A., Ostojic, S., and Pillow, J.
\newblock Probing the relationship between linear dynamical systems and
  low-rank recurrent neural network models.
\newblock \emph{arXiv preprint arXiv:2110.09804}, 2021.

\bibitem[Von~Oswald et~al.(2023)Von~Oswald, Niklasson, Randazzo, Sacramento,
  Mordvintsev, Zhmoginov, and Vladymyrov]{von2023transformers}
Von~Oswald, J., Niklasson, E., Randazzo, E., Sacramento, J., Mordvintsev, A.,
  Zhmoginov, A., and Vladymyrov, M.
\newblock Transformers learn in-context by gradient descent.
\newblock In \emph{International Conference on Machine Learning}, pp.\
  35151--35174. PMLR, 2023.

\bibitem[Vyas et~al.(2020)Vyas, Golub, Sussillo, and
  Shenoy]{vyas2020computation}
Vyas, S., Golub, M.~D., Sussillo, D., and Shenoy, K.~V.
\newblock Computation through neural population dynamics.
\newblock \emph{Annual Review of Neuroscience}, 43:\penalty0 249--275, 2020.

\bibitem[Wang et~al.(2024)Wang, Dong, Jiang, Ji, Liu, and
  Wu]{wang2024brainscale}
Wang, C., Dong, X., Jiang, J., Ji, Z., Liu, X., and Wu, S.
\newblock Brainscale: Enabling scalable online learning in spiking neural
  networks.
\newblock \emph{bioRxiv}, pp.\  2024--09, 2024.

\bibitem[Werfel et~al.(2003)Werfel, Xie, and Seung]{werfel2003learning}
Werfel, J., Xie, X., and Seung, H.
\newblock Learning curves for stochastic gradient descent in linear feedforward
  networks.
\newblock \emph{Advances in neural information processing systems}, 16, 2003.

\bibitem[Williams et~al.(2021)Williams, Kunz, Kornblith, and
  Linderman]{williams2021generalized}
Williams, A.~H., Kunz, E., Kornblith, S., and Linderman, S.
\newblock Generalized shape metrics on neural representations.
\newblock \emph{Advances in Neural Information Processing Systems},
  34:\penalty0 4738--4750, 2021.

\bibitem[Woodworth et~al.(2020)Woodworth, Gunasekar, Lee, Moroshko, Savarese,
  Golan, Soudry, and Srebro]{woodworth2020kernel}
Woodworth, B., Gunasekar, S., Lee, J.~D., Moroshko, E., Savarese, P., Golan,
  I., Soudry, D., and Srebro, N.
\newblock Kernel and rich regimes in overparametrized models.
\newblock In \emph{Conference on Learning Theory}, pp.\  3635--3673. PMLR,
  2020.

\bibitem[Xie \& Seung(1999)Xie and Seung]{xie1999spike}
Xie, X. and Seung, H.~S.
\newblock Spike-based learning rules and stabilization of persistent neural
  activity.
\newblock \emph{Advances in neural information processing systems}, 12, 1999.

\bibitem[Yamins et~al.(2014)Yamins, Hong, Cadieu, Solomon, Seibert, and
  DiCarlo]{yamins2014performance}
Yamins, D.~L., Hong, H., Cadieu, C.~F., Solomon, E.~A., Seibert, D., and
  DiCarlo, J.~J.
\newblock Performance-optimized hierarchical models predict neural responses in
  higher visual cortex.
\newblock \emph{Proceedings of the national academy of sciences}, 111\penalty0
  (23):\penalty0 8619--8624, 2014.

\bibitem[Yang \& Molano-Mazón(2021)Yang and Molano-Mazón]{yang2021next}
Yang, G.~R. and Molano-Mazón, M.
\newblock Towards the next generation of recurrent network models for cognitive
  neuroscience.
\newblock \emph{Current Opinion in Neurobiology}, 70:\penalty0 182--192, 2021.

\bibitem[Yang \& Wang(2020)Yang and Wang]{yang2020artificial}
Yang, G.~R. and Wang, X.-J.
\newblock Artificial neural networks for neuroscientists: a primer.
\newblock \emph{Neuron}, 107\penalty0 (6):\penalty0 1048--1070, 2020.

\bibitem[Yang et~al.(2019)Yang, Joglekar, Song, Newsome, and
  Wang]{yang2019task}
Yang, G.~R., Joglekar, M.~R., Song, H.~F., Newsome, W.~T., and Wang, X.-J.
\newblock Task representations in neural networks trained to perform many
  cognitive tasks.
\newblock \emph{Nature neuroscience}, 22\penalty0 (2):\penalty0 297--306, 2019.

\bibitem[Zador(2019)]{zador2019critique}
Zador, A.~M.
\newblock A critique of pure learning and what artificial neural networks can
  learn from animal brains.
\newblock \emph{Nature communications}, 10\penalty0 (1):\penalty0 1--7, 2019.

\bibitem[Zahorodnii et~al.(2025)Zahorodnii, Mendoza-Halliday,
  Martinez-Trujillo, Qian, Desimone, and Cueva]{zahorodnii2025overcoming}
Zahorodnii, A., Mendoza-Halliday, D., Martinez-Trujillo, J.~C., Qian, N.,
  Desimone, R., and Cueva, C.~J.
\newblock Overcoming sensory-memory interference in working memory circuits.
\newblock \emph{bioRxiv}, pp.\  2025--03, 2025.

\bibitem[Zipser(1991)]{Zipser1991}
Zipser, D.
\newblock Recurrent network model of the neural mechanism of short-term active
  memory.
\newblock \emph{Neural Computation}, 3\penalty0 (2):\penalty0 179--193, 1991.

\end{thebibliography}
\bibliographystyle{icml2025}

\newpage
\appendix
\onecolumn

\section{Methods} \label{scn:methods} 

\subsection{Further Details on the Neural Datasets and Synthetic Data for RNN Training} 

The \textbf{Mante 2013} dataset was downloaded from \url{https://www.ini.uzh.ch/en/research/groups/mante/data.html}. We trained RNNs using a synthetic task setup from Neurogym~\cite{molano2022neurogym}, which included a 350 ms fixation period, a 750 ms stimulus presentation period, a 300 ms delay period, and a 300 ms decision period. The activity of the trained RNNs during the stimulus period was then compared to the downloaded neural dataset using the aforementioned similarity measures. A grid search on the fixation and decision interval durations revealed only minor differences in distances and a consistent trend across learning rules.

The \textbf{Sussillo 2015} dataset consisted of electrode recordings from primary motor (M1) and dorsal premotor cortex (PMd) taken while a monkey performed a maze-reaching task consisting of 27 differerent reaching conditions~\cite{sussillo2015neural}. To assess the similarity between the neural activity and RNNs we compared activity from -1450 ms to 400 ms relative to movement onset. The inputs and outputs to train the models were described in Sussillo et al. 2015, but in brief, for each reach condition there were 16 inputs and 7 target outputs. The 7 outputs were the electromyographic (EMG) signals recorded from 7 muscles as the monkey performed a reaching movement. 15 inputs specified the upcoming reach condition, and were derived from preparatory period neural activity. The remaining input was a hold-cue that took a value of +1 before movemement onset and then a value of 0 to initiate the movement, whereupon the model generated the 7 EMG signals. 

\subsection{Further Details on the Learning Rule} \label{scn:learning_rules}

This subsection aims to clarify the approximation mechanisms employed by each bio-plausible learning rule. For comprehensive descriptions, we recommend consulting the detailed references provided. We begin by expressing the gradient via real-time recurrent learning (RTRL) factorization (an equivalent but causal alternative to the BPTT factorization of the gradient):
\begin{eqnarray}
    \frac{d L}{d W_{h, ij}} &=&\sum_{l,t}\frac{\partial L}{\partial h_{l,t}} \frac{d h_{l,t}}{d W_{h, ij}}, \label{eqn:sum_tl} 
\end{eqnarray}

We follow the total versus partial derivative notation (\(d\) vs.\ \(\partial\)) as in~\cite{bellec2020solution}. The primary challenge with RTRL in terms of biological plausibility and computational efficiency lies in the term $\frac{d h_{l,t}}{d W_{h, ij}}$ from the gradient decomposition (Eq.~\ref{eqn:sum_tl}). This term tracks all recursive dependencies of $h_{l,t}$ on the weight $W_{h,ij}$ due to recurrent connections, calculated recursively as:
\begin{align}
    \frac{d h_{l,t}}{d W_{h, ij}} &= \frac{\partial h_{j,t}}{\partial W_{h,ij}} + \sum_m \frac{\partial h_{l,t}}{\partial h_{m,t-1}} \frac{d h_{m,t-1}}{d W_{h,ij}} \cr 
    &= \frac{\partial h_{l,t}}{\partial W_{h,ij}} + \frac{\partial h_{l,t}}{\partial h_{l,t-1}} \frac{d h_{l,t-1}}{d W_{h,ij}} + 
    {\underbrace{\textstyle \sum_{m\neq l} W_{h,lm} f'(h_{m,t-1}) \frac{d h_{m,t-1}}{d W_{h,ij}} }_{\mathclap{ \text{\normalsize involving all weights $W_{h,lm}$}}}}. \label{eqn:s_triple}
\end{align}
Consequently, $\frac{d h_{l,t}}{d W_{h, ij}}$ presents a significant challenge for biological plausibility as it includes nonlocal terms, necessitating knowledge of all other network weights for updating each $W_{h,ij}$. \textbf{For a learning rule to be biologically plausible, all information required to update a synaptic weight must be physically accessible to that synapse. However, it remains unclear how neural circuits could make such extensive information readily available to every synapse.}  

Approaches like \textbf{e-prop}~\cite{bellec2020solution} and equivalently, \textbf{RFLO}~\cite{murray2019local}, address this by truncating the problematic nonlocal terms in Eq.~\ref{eqn:s_triple}, ensuring that updates to $W_{h,ij}$ follow a three-factor framework --- the updates rely solely on local pre- and post-synaptic activity and a third top-down instructive signal (e.g. from neuromodulators):
\begin{align}
    \widehat{\frac{\partial h_{l,t}}{\partial W_{h, ij}}} &= \begin{cases} 
    \frac{\partial h_{i,t}}{\partial W_{h,ij}} + \frac{\partial h_{i,t}}{\partial h_{i,t-1}} \widehat{\frac{\partial h_{i,t-1}}{\partial W_{h,ij}}},  & l = i \\
    0, & l \neq i
    \end{cases}
\end{align}
which yields a much simpler factor than the comprehensive tensor depicted in Eq.~\ref{eqn:s_triple}. This truncation can be achieved in PyTorch using $h.detach()$, preventing gradient propagation through the recurrent weights. 

Putting this together, e-prop can be written in terms of known biological processes including --- eligibility trace $e$ and top-down instructive signals $I$ --- as~\cite{bellec2020solution}: 
\begin{equation}
    \left. \Delta W_{h,ij} \right|_{e-prop} =  \sum_{t} I_{i,t} e_{ij,t},
\end{equation}
where $I_{i,t} = \frac{\partial L}{\partial h_{i,t}}$ is the top-down instructive signal (e.g. from neuromodulator dopamine, neuronal firing, etc.~\cite{Gerstner2018,bellec2020solution}) sent to neuron $i$ at time $t$, and $e_{ij,t}=\widehat{\frac{\partial h_{i,t}}{\partial W_{h, ij}}}=\frac{\partial h_{i,t}}{\partial W_{h,ij}} + \frac{\partial h_{i,t}}{\partial h_{i,t-1}} \widehat{\frac{\partial h_{i,t-1}}{\partial W_{h, ij}}}$ is the eligibility trace for synapse $(ij)$ at time $t$. This is a three-factor rule, with the pre-and postsynaptic neuron factors in the eligibility trace as well as a third factor from the instructive signal. 

Besides eligibility traces and top-down instructive signals, recent transcriptomics data~\cite{Smith2020} suggest the presence of widespread cell-type-specific local modulatory signals that could convey additional information for guiding synaptic weight updates. \textbf{ModProp} is developed to incorporate these processes and restore the gradient terms truncated by e-prop, thereby improving the approximation of the gradient. Specifically, the ModProp update rule is described as follows~\cite{liu2022biologically}:
\begin{align}
    \left. \Delta W_{h,ij} \right|_{ModProp}
    &\propto I_i \times e_{ij} + \left( \sum_{\alpha \in C} \left( \sum_{l\in \alpha} I_l h'_l \right) \times F_{\alpha \beta}\right) * e_{ij}, \cr
    F_{\alpha \beta, s} &= \mu^{s-1} (W^s)_{\alpha \beta}, \label{eq_ModProp}
\end{align}
where $I$ and $e$ again denote the top-down learning signal and the eligibility trace, respectively. Here, neuron $j$ belongs to type $\alpha$, neuron $p$ to type $\beta$, and $C$ denotes the set of cell types. $F_{\alpha \beta}$ is hypothesized to represent type-specific filter taps of GPCRs expressed by cells of type $\beta$ in response to precursors secreted by cells of type $\alpha$. The operator $*$ denotes convolution, and $s$ indexes the filter taps. The hyperparameter $\mu$, set to 0.25 in this study, and the genetically predetermined $(W^s)_{\alpha \beta}$ values for different filter taps $F_{\alpha \beta, s}$ could be optimized over evolutionary timescales~\cite{liu2022biologically}.

We also explored an older learning rule, \textbf{node perturbation}~\cite{werfel2003learning,lillicrap2016random}, which is known to have trouble scaling beyond small-scale networks and simple tasks. Specifically, it is implemented by 
\begin{equation}
    \left. \Delta W_{h,ij} \right|_{NP} \propto  \sum_{t} \widehat{I_{i,t}} e_{ij,t},
\end{equation}
where $\widehat{I_{t}} = (L_t(h_t+\xi)-L_t(h_t))\xi/\sigma^2$ provides an estimate to $\frac{\partial L}{\partial h_{t}}$; elements of $\xi$ are chosen independently from a zero-mean Gaussian distribution with variance $\sigma^2$.

In addition, we explored \textbf{evolutionary strategies}~\cite{salimans2017evolution} for parameter updates in our model. This method, for a Gaussian distribution, is implemented as follows:
\begin{equation}
\left. \Delta W_{h,ij} \right|_{ES} \propto \frac{1}{\sigma S} \sum_{s=1}^S L^{(s)} \epsilon^{(s)},
\end{equation}
where $\epsilon^{(s)}$ is sampled from a standard normal distribution $\mathcal{N}(0,I)$ for $s=1,...,S$. Here, $L^{(s)}$ represents the loss function evaluated after perturbing the parameter by $\sigma \epsilon^{(s)}$, $\sigma$ is the standard deviation of the perturbations, and $S$ is the number of samples. Due to computational constraints, we set $S$ to $50$ for our experiments.

\subsection{Additional Details on Training and Analysis} \label{scn:sim_details}

Our model-data comparison method utilizes Procrustes distances, as implemented in \url{https://github.com/ahwillia/netrep}, with the configuration set to $metric = LinearMetric(alpha=1.0, center\_columns=True)$. Additionally, in Appendix Figure~\ref{fig:W_eig}, we employed Dynamical Systems Analysis (DSA), available at \url{https://github.com/mitchellostrow/DSA/tree/main}. For this analysis, we tested with hyperparameters $n\_delays \in \{5, 10, 15, 20\}$ and $rank \in \{10, 20, 30, 40\}$, observing consistent trends across settings. We did not test values beyond these ranges due to computational resource limitations. For the loss used in training RNNs, we used cross-entropy loss for the Mante 2013 task and mean-squared error for the Sussillo 2015 task (with EMG outputs as the targets~\cite{cloos2022scaling}). As mentioned, the Mante 2013 dataset was downloaded from \url{https://www.ini.uzh.ch/en/research/groups/mante/data.html}. The Hatsopoulos 2007 dataset was downloaded from \url{https://datadryad.org/dataset/doi:10.5061/dryad.xsj3tx9cm}. However, we obtained the Sussillo 2015 dataset from the original authors and do not have permission to redistribute it. 

Our code is available at \url{https://github.com/Helena-Yuhan-Liu/LearningRuleSimilarities/tree/main}. Both YHL and CJC contributed to the code development. We utilized PyTorch Version 1.10.2 \citep{paszke2019pytorch}. Simulations were executed on a server equipped with two 20-core Intel(R) Xeon(R) CPU E5-2698 v4 at 2.20GHz. The average training duration for tasks was about 10 minutes, and the analysis pipeline required approximately 2 minutes per model. Training employed the Adam optimizer. Unless otherwise noted, the learning rate was set at $1e-3$, optimized through a grid search of $\{3e-3, 1e-3, 3e-4, 1e-4\}$. We used a batch size of around $100$; changes in this parameter led to negligible differences in the results. The number of time steps, $T$,  for the Sussillo task was set to $186$, matching the original data. The number of time steps $T$, for the Mante task was $34$, based on $dt=50$ ms from the original Mante paper and the total task duration in the Neurogym setting. Similar trends were observed when we varied dt and the durations of the fixation and delay periods. We employed $200$ hidden units for the Sussillo 2015 task and $400$ hidden units for the Mante 2013 task; doubling these numbers resulted in similar trends. Each simulation was repeated with four different seeds (except for 10 seeds for Figure~\ref{fig:lrsim}B), and results for each seed were plotted as separate lines in our figures. Training involved 1000 SGD iterations for Sussillo 2015 and 3000 for Mante 2013, with input, recurrent, and readout weights all trainable. Local learning rule approximations were specifically applied to input and recurrent weights, due to the locality issues discussed in Section~\ref{scn:learning_rules}. Unlike these weights, readout weights do not encounter such issues; hence, by default, the same readout weights were used for both forward and backward computations. However, as verified in Appendix Figure~\ref{fig:FA}, employing random feedback readout weights for training (i.e., feedback alignment~\cite{lillicrap2016random}) resulted in comparable distances. 

By default, zero-mean Gaussian noise with a standard deviation of $0.1$ was added to the hidden activity, except in cases where the noise was removed to assess its impact. Typically, no connectivity constraints were applied, except for settings in Figure~\ref{fig:var_cond}B where $80\%$ of the neurons were enforced as strictly excitatory and $20\%$ as inhibitory. To enforce Dale's law, we used the same masking procedure in ~\cite{yang2020artificial}. Additionally, the Sussillo 2015 plot in Figure~\ref{fig:default_distances} applies a sparsity constraint, limiting only $25\%$ of the recurrent weights to be nonzero and trainable. To initialize the weights, we initialized with random Gaussian distributions where each weight element \(W_{h,ij} \sim \mathcal{N}(0, g^2/N)\), with an initial weight variance of \(g\); unless otherwise mentioned, we set \(g=1.0\). Input and readout weights were initialized similarly as in \cite{yang2020artificial} (see their \(EIRNN.ipynb\) notebook).

Normalized accuracy, which appears as the x-axis in several plots, is defined such that a value of $1$ corresponds to perfect performance. For Sussillo 2015, normalized accuracy is calculated as \(1 - \text{normalized mean squared error}\), as used in~\cite{depasquale2018full}. In the case of Mante 2013, which involves a classification task where mean squared error is not applicable, normalized accuracy is computed as \(1 - \text{cross entropy loss}\) to maintain consistency with the definition where $1$ indicates the best performance. We also applied x-axis limits to constrain the range between 0 and 1 for uniformity. While we match task accuracies for neural similarity comparisons, we note that --- as observed in prior work~\cite{bellec2020solution} --- e-prop typically requires more training iterations than BPTT to reach the same accuracy. 

We detail the data-splitting procedure used for generating the noise floor, i.e. the baseline, in Figure~\ref{fig:noise_floor}. We split the neural data into nonoverlapping groups each containing $N_{sample}$ neurons ($ineurons1$, $ineurons2$). We sample $N_{sample}$ units from the RNN model ($iunits$). We compute the distance between two samples of neural data $d1 = D(ineurons2, ineurons1)$. $d1$ is the lowest we can hope to get given the variability in the neurons that were recorded. We compute the distance between samples of the model and neural data $d2 = D(iunits, ineurons1)$. For each iteration of this procedure we get a new estimate for the distance between the model and data, and the data-to-data distance. 

\begin{table*}[ht]
\centering
\label{tab:trial_averaging}
\begin{tabular}{lcccccc}
\toprule
\textbf{Number of trials averaged} & \textbf{1} & \textbf{5} & \textbf{10} & \textbf{15} & \textbf{20} \\
\midrule
Distance & 0.513 ± 0.008 & 0.358 ± 0.005 & 0.280 ± 0.004 & 0.238 ± 0.003 & 0.212 ± 0.003 \\
\bottomrule
\end{tabular}
\caption{Illustrating greater data-data distance when computed from single trials; note that these numbers are not comparable to those in Figure~\ref{fig:noise_floor} due to different ways of subsampling.}
\end{table*}

Although the neural datasets used in this study contain multiple trials per condition, we did not compute data-data similarity via trial-based subsampling. This decision reflects the substantial trial-to-trial variability in single-trial firing rate estimates, which can dominate the distance metric and obscure meaningful similarities—even between similar neural responses~\cite{cunningham2009methods,kay2024disentangling}. As illustrated in Table~\ref{tab:trial_averaging}, data-data distances are substantially higher when computed from single trials, but decrease with trial averaging. This trend reflects how averaging across trials helps recover stable, condition-specific firing patterns. While existing methods (e.g., LFADS~\cite{pandarinath2018inferring}) have improved single-trial inference, accurate estimation of latent dynamics from single trials remains an open challenge. Therefore, following standard practice in systems neuroscience, we computed similarity based on trial-averaged responses. Our chosen data-data baseline (used in Figure~\ref{fig:noise_floor}) compares held-out subsets of neurons recorded under matched task conditions and timepoints, and serves to assess a practical question: if more neurons had been recorded from the same brain region, would they be distinguishable from model units? 

\newpage 

\section{Convergence and Divergence of E-prop Updates in a Toy Setting} \label{scn:proof} 

We present the formal statement and proof for Proposition~\ref{thm:eprop_converge} in Proposition~
\ref{prop:converge_detailed}. 

Consider a 1-dimensional linear recurrent neural network (RNN) with a single scalar recurrent weight parameter \( W \), input sequence \( \{x_t\}_{t=1}^T \subset \mathbb{R} \), and output \( \hat{y} \) defined as the readout of the last hidden state \( h_T \): 
\begin{equation}
    h_{t} = W h_{t-1} + x_t, \quad \hat{y} = h_T, \label{eqn:linearRNN_a}
\end{equation}
leading to
\begin{equation}
\hat{y}(W) = \sum_{t=0}^{T-1} W^t x_{T-t}. \label{eqn:linearRNN_b}
\end{equation}
The loss function is the least-squares loss \( L = \frac{1}{2} (\hat{y} - y)^2 \). Assume a target \( y = 0 \), making \( L = \frac{1}{2} \hat{y}^2 \). We note that this simple setup effectively reduces to a polynomial root finding problem, with \( \{x_t\}_{t=1}^T \subset \mathbb{R} \) as the coefficients. The corresponding gradient-based updates are modeled as a continuous-time dynamical system (CTDS), with specific forms for BPTT and e-prop updates:
\begin{equation}
\tau \frac{dW(t)}{dt} \Big|_\text{BPTT} = - \frac{dL}{dW} = -\hat{y}(W) \hat{y}'(W), \quad \tau \frac{dW(t)}{dt} \Big|_\text{e-prop} = - \hat{y}(W) x_{T-1}. \label{eqn:linearRNN_c}
\end{equation}

\begin{prop} \label{prop:converge_detailed}

Consider the setup in Eqs.~\ref{eqn:linearRNN_a}-\ref{eqn:linearRNN_c} and assume the polynomial \( \hat{y}(W) \) has at least one real root \( W^* \), where \( \hat{y}(W^*) = 0 \), and BPTT converges to \( W^* \); the Jacobian \( J_\text{e-prop}(W) \) evaluated at \( W^* \) satisfies \( \text{Re}(\lambda) < 0 \), where \( \lambda \) is the eigenvalue of \( J_\text{e-prop}(W^*) \); the input sequence \( \{x_t\}_{t=1}^T \) satisfies \( x_{T-1} \neq 0 \), ensuring non-degenerate e-prop updates. 

If \( W(0) \) satisfies \( \|W(0) - W^*\| < \delta \), where \( \delta > 0 \) is the radius of a neighborhood around \( W^* \) within which \( \text{Re}(\lambda) \text{ of } J_\text{e-prop}(W) < 0 \) for all \( W \), requiring \( \text{sign}(\hat{y}'(W)) = \text{sign}(x_{T-1}) \) upon initialization, then the e-prop update will asymptotically converge to \( W^* \). If \( W(0) \) is initialized such that \( \text{sign}(\hat{y}'(W)) = - \text{sign}(x_{T-1}) \)  and $\text{sign}(\hat{y}'(W(t))$ remains constant for $t\geq 0$, then e-prop will diverge. 

\end{prop}

\begin{proof} 
We analyze the system governed by the e-prop updates as a continuous-time dynamical system (CTDS):
\[
\tau \frac{dW(t)}{dt} = - \hat{y}(W) x_{T-1},
\]
where the Jacobian is:
\[
J_\text{e-prop}(W) = \frac{\partial}{\partial W} \left( \frac{dW(t)}{dt} \right) = -\frac{1}{\tau} \hat{y}'(W) x_{T-1}.
\]

By definition, since $J_\text{e-prop}(W^*)$ is a scalar for 1-dimensional RNN, its eigenvalue is itself, and \( W^* \) is an asymptotically stable fixed point if and only if \( J_\text{e-prop}(W^*) < 0 \), ensuring that trajectories initialized sufficiently close to \( W^* \) converge to \( W^* \). 

We start by proving the existence of a case for convergence. Evaluate the Jacobian at \( W^* \):
\[
J_\text{e-prop}(W^*) = - \frac{1}{\tau} \hat{y}'(W^*) x_{T-1}.
\]
Having \( J_\text{e-prop}(W^*) < 0 \) implies:
\[
\hat{y}'(W^*) x_{T-1} > 0.
\]

As provided in the Proposition statement, suppose \( W(0) \) is initialized such that \( \|W(0) - W^*\| < \delta \) for some \( \delta > 0 \) (within the asymptotic stability region of \( W^* \)), within which the Jacobian remains negative \( J_\text{e-prop}(W(t)) < 0 \quad \forall t \geq 0\). In this region, the CTDS is a contracting flow:
\[
\frac{d}{dt} \|W(t) - W^*\|^2 = 2 (W(t) - W^*) \frac{dW(t)}{dt}.
\]
Substitute the update rule:
\[
\frac{d}{dt} \|W(t) - W^*\|^2 = -2 \frac{1}{\tau} (W(t) - W^*) \hat{y}(W(t)) x_{T-1}.
\]
Using Taylor expansion near \( W^* \), \( \hat{y}(W(t)) \approx \hat{y}'(W^*) (W(t) - W^*) \). Thus:
\[
\frac{d}{dt} \|W(t) - W^*\|^2 = -2 \frac{1}{\tau} \hat{y}'(W^*) x_{T-1} (W(t) - W^*)^2.
\]
Since \( \hat{y}'(W^*) x_{T-1} > 0 \), this ensures \( \|W(t) - W^*\|^2 \) decreases monotonically, proving convergence to \( W^* \). As a side note, since $W(t)$ and $W^*$ are scalars, \( \|W(t) - W^*\|^2 = (W(t) - W^*)^2\).   

We next prove the existence of a case for failure to converge. By the Proposition statement, if \( W(0) \) is initialized such that \( \text{sign}(\hat{y}'(W)) = - \text{sign}(x_{T-1}) \),  $\text{sign}(\hat{y}'(W(t))$ remains constant for $t\geq 0$ and $x_{T-1}$ is a constant, then
\[
\hat{y}'(W)) x_{T-1} < 0 \quad \forall t \geq 0.
\]

This would imply that 
\[
J_\text{e-prop}(W(t)) > 0 \quad \forall t \geq 0.
\]

Since the eigenvalue of Jacobian \( J_\text{e-prop}(W(t)) \) (i.e. itself, since $J$ is a scalar here) has strictly positive real part for all \( t \geq 0 \) along the trajectory \( W(t) \), \( W(t) \) does not converge to any stable fixed point, including \( W^* \). 

\end{proof}

To help visualize this theoretical result (facilitated by the simplicity of the polynomial root-finding setup), consider the following scenarios. For stable initialization, requiring the initialization to satisfy \( \text{sign}(\hat{y}'(W)) = \text{sign}(x_{T-1}) \) would mean that the trajectory begins on a rising edge when \( \text{sign}(x_{T-1}) > 0 \) or a falling edge when \( \text{sign}(x_{T-1}) < 0 \). For the unstable initialization case, requiring the initialization to satisfy \( \text{sign}(\hat{y}'(W)) \neq \text{sign}(x_{T-1}) \) would mean that the trajectory starts on an edge with mismatched signs, leading to initial divergence. However, if \( \text{sign}(\hat{y}'(W(t))) \) changes during the trajectory (e.g., by entering a different region), the system may settle into a different basin of attraction; this scenario would violate the assumption of a constant \( \text{sign}(\hat{y}'(W(t))) \) for all \( t \geq 0 \). A particular case of instability occurs when \( \text{sign}(x_{T-1}) < 0 \) at initialization and the trajectory is initialized on a far-right (or far-left) rising edge. Here, the weight updates diverge without entering any basin. 

\newpage 

\section{Additional Simulations}

In Appendix Figure~\ref{fig:pairDist_embedding_snapshot} displays the similarity among models in terms of their pairwise distances and their embeddings across different sampled training snapshots. In Appendix Figure~\ref{fig:var_cond}, we demonstrate consistent patterns when recurrent noise is removed and Dale's law is enforced. We also explore ModProp~\cite{liu2022biologically}, which incorporates cell-type-specific local modulatory signals to reintroduce terms omitted by e-prop; however, as ModProp is effective only under specific conditions (Dale's law and $ReLU$ activation), confining Appendix Figure~\ref{fig:var_cond}B to these settings. Further analysis of post-training weight eigenspectrums and distances, conducted using Dynamical Similarity Analysis (DSA), reinforces the similarity between BPTT and e-prop, as shown in Appendix Figure~\ref{fig:W_eig}.

\newpage 

\begin{figure*}[h!]
    \centering
    \includegraphics[width=0.9\textwidth]{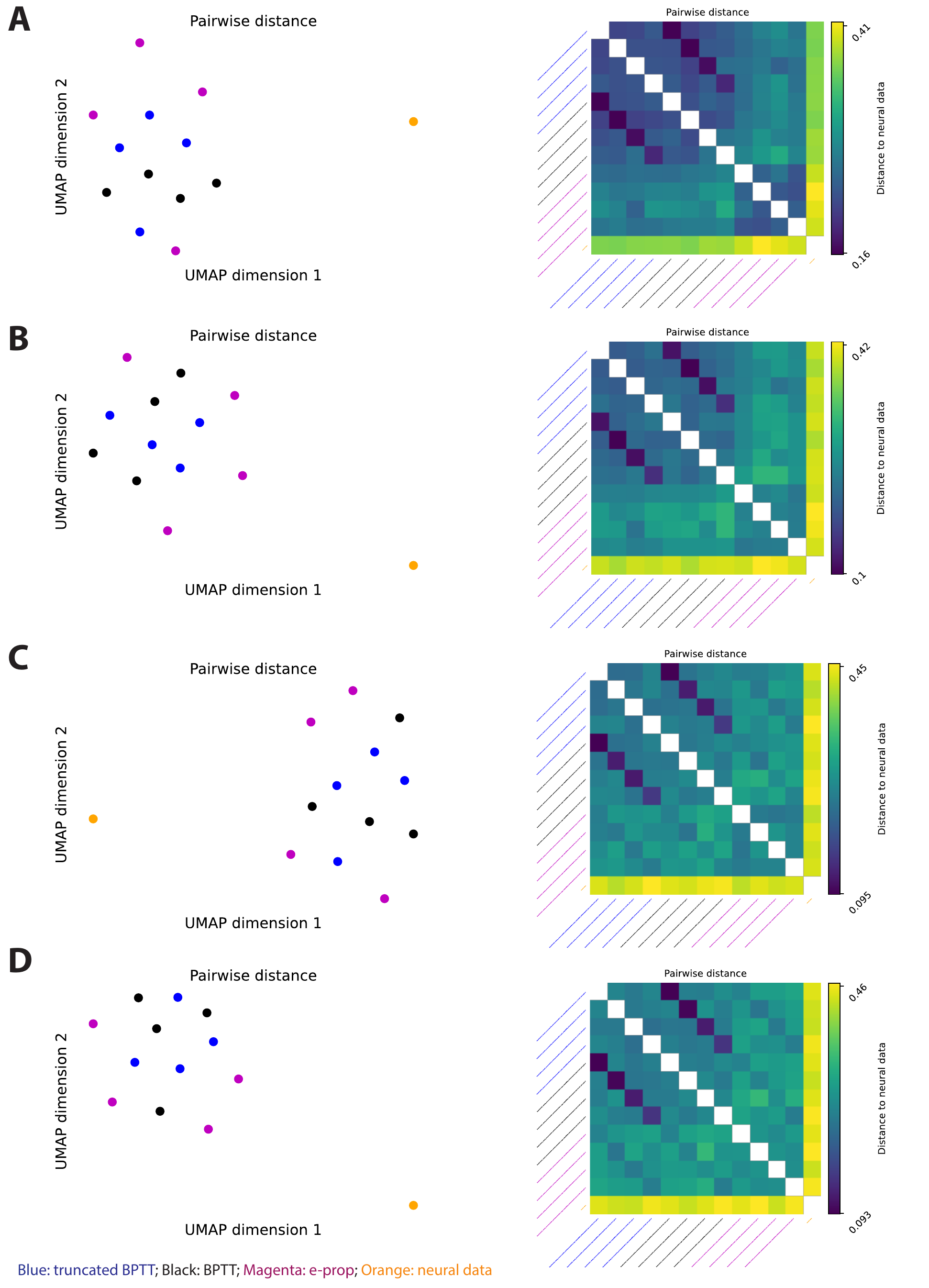}
    \caption{ UMAP embedding and pairwise distance matrix heatmap for different models when (A) 
 best e-prop accuracy, (B) $80\%$, (C) $60\%$, and (D) $40\%$ accuracies are reached. Here, the Sussillo 2015 dataset is illustrated. Black: BPTT, blue: truncated BPTT, magenta: e-prop, orange: neural data. The pairwise distances show similarities across learning rules relative to data, indicated by lower distances between models as compared to model-data distance. 
    } 
    \label{fig:pairDist_embedding_snapshot}
\end{figure*} 

\newpage 

\begin{figure*}[h!]
    \centering
    \includegraphics[width=0.99\textwidth]{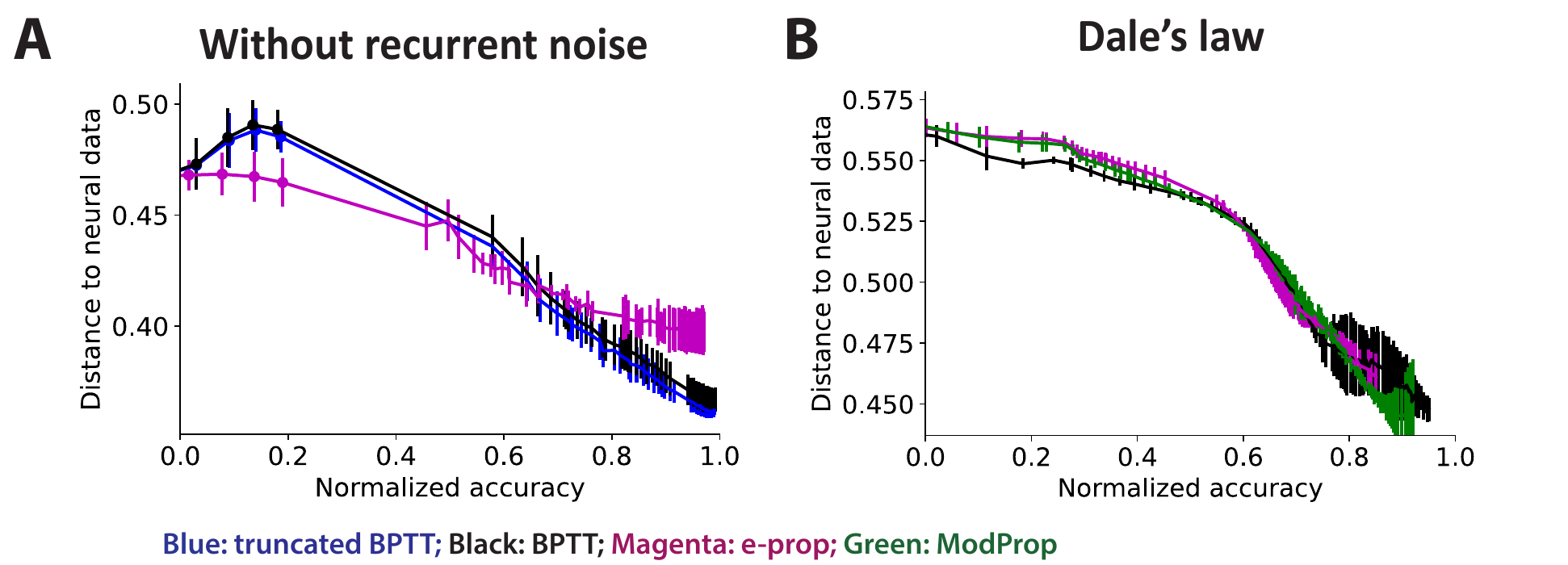}
    \caption{ \textbf{This figure replicates the core trend from Figure~\ref{fig:default_distances} --- namely, comparable similarities across rules --- under different settings.} The settings include: (A) removal of RNN hidden activity noise, and (B) enforcement of Dale’s law. This plot uses the Sussillo2015 dataset as an illustrative example and compares Procrustes distances versus accuracy for three learning rules: BPTT (black), e-prop (magenta), and ModProp (green) --- the latter functioning exclusively under Dale’s law constraint and $ReLU$ activation. Plotting conventions mirroring those in Figure~\ref{fig:default_distances}. 
    } 
    \label{fig:var_cond}
\end{figure*} 

\newpage

\begin{figure*}[h!]
    \centering
    \includegraphics[width=0.99\textwidth]{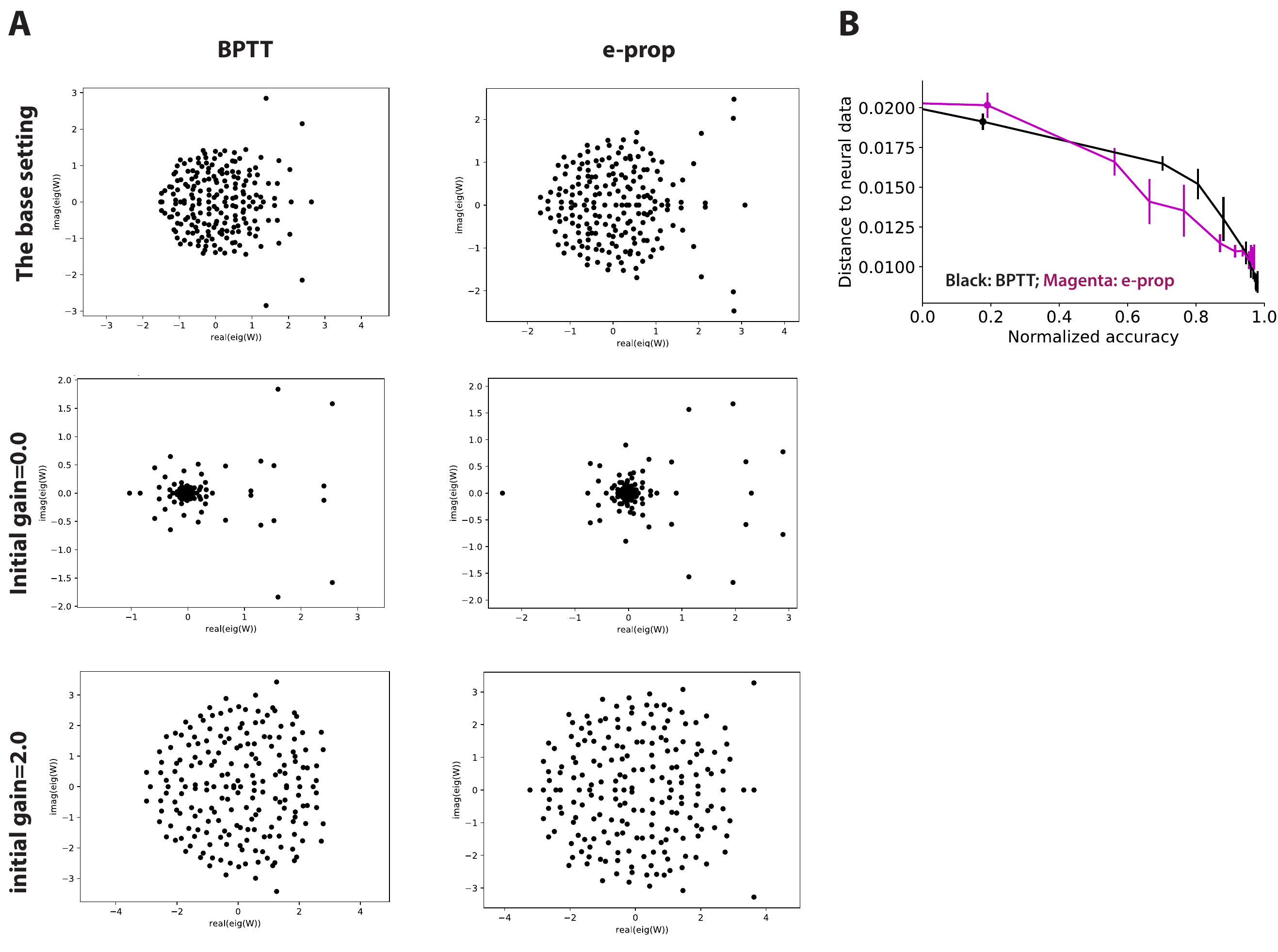}
    \caption{ (A) presents the eigenvalues of the recurrent weight matrix post-training, with columns representing BPTT and e-prop respectively. Each row displays a different training setting: the base setting (referenced in Figure~\ref{fig:default_distances}), initial weight standard deviation set to 0, and initial weight standard deviation set to $2/\sqrt{N}$. Notably, eigenvalue distributions appear more similar within each setting across learning rules (BPTT vs. e-prop) than across different settings for the same learning rule, further highlighting the similarity between BPTT and e-prop. B) The Dynamical Similarity Analysis (DSA), which evaluates systems based on their dynamical characteristics, is also unable to distinguish between learning rules when considering their proximity to neural data. Similar to previous figures, e-prop is plotted in magenta and BPTT is plotted in black.   
    } 
    \label{fig:W_eig}
\end{figure*} 

\newpage 

\begin{figure*}[h!]
    \centering
    \includegraphics[width=0.99\textwidth]{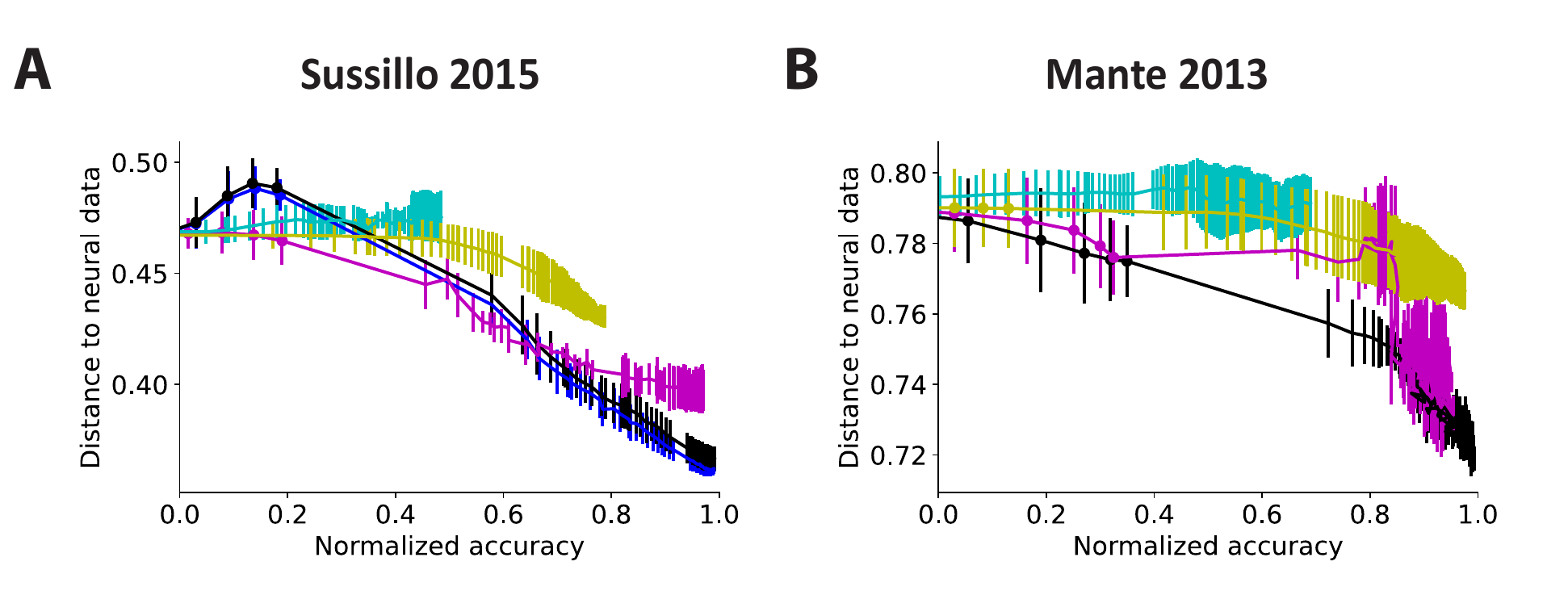}
    \caption{ Node perturbation (cyan) and evolutionary strategies (yellow) lead to higher Procrustes distances from the neural data compared to BPTT (black) and e-prop (magenta) when accuracies are equivalent. This figure presents the Procrustes distance versus accuracy plots, adhering to the plotting conventions established in Figure~\ref{fig:default_distances}, for (A) the Sussillo 2015 task and (B) the Mante 2013 task. Here, simulations are done without recurrent noise, as in Appendix Figure~\ref{fig:var_cond}A, for a more stable performance of some learning rules. 
    } 
    \label{fig:nodeP}
\end{figure*} 

\begin{figure*}[h!]
    \centering
    \includegraphics[width=0.99\textwidth]{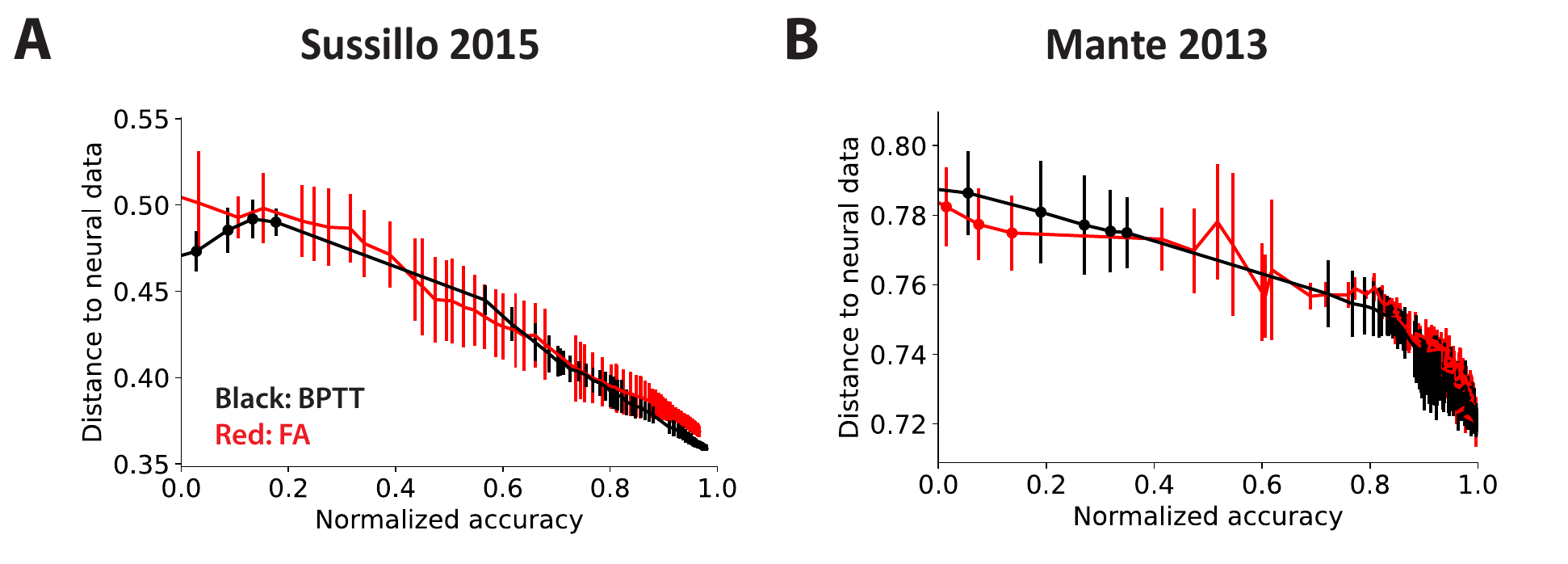}
    \caption{ The use of random feedback readout weights for gradient computation (red)~\cite{lillicrap2016random} resulted in distances comparable to those achieved using exact readout weights (black). Plotting conventions are consistent with previous figures: task performance varies with training iterations, and error bars denote variation across seeds. 
    } 
    \label{fig:FA}
\end{figure*} 

\end{document}